\documentclass[a4paper,fleqn]{cas-sc}

\usepackage[numbers]{natbib}

\usepackage{lscape} 
\usepackage{amssymb}

\usepackage{amsfonts}

\usepackage[english]{babel}

\usepackage{amsmath}
\usepackage{graphicx}
\usepackage{subcaption}
\usepackage{rotating}
\usepackage{pifont}
\usepackage{color}

\def\tsc#1{\csdef{#1}{\textsc{\lowercase{#1}}\xspace}}
\tsc{WGM}
\tsc{QE}

\begin{document}
\let\WriteBookmarks\relax
\def\floatpagepagefraction{1}
\def\textpagefraction{.001}

\shorttitle{PSGformer: Enhancing 3D Point Cloud Instance Segmentation via Precise Semantic Guidance}    

\shortauthors{Wuyang Luan et al.}  

\title [mode = title]{PSGformer: Enhancing 3D Point Cloud Instance Segmentation via Precise Semantic Guidance}  

\author[1]{Lei Pan}
\ead{leipan.cafuc@hotmail.com}
\credit{supervision, writing}
\fnmark[1]
\fntext[1]{Lei Pan received his PhD from UESTC University }

\author[1]{Wuyang Luan}
\ead{lwy3137384@gmail.com}
\credit{Conceptualization of this study, Methodology, Software}
\fnmark[2]
\fntext[2]{Wuyang Luan is a student in  China Civil
Aviation Flight Academy of China University}

\author[1]{Yuan Zheng}
\ead{ranchozy@cafuc.edu.cn}
\credit{supervision, writing}
\fnmark[3]
\fntext[2]{Yuan Zheng received his PhD from Zhejiang University}

\author[1]{Qiang Fu}
\ead{cafczdm@163.com}
\credit{supervision, writing}
\fnmark[4]
\fntext[4]{Qiang Fu is the dean of School of Computer Science, China Civil Aviation Flight Academy of China}

\author[1]{Junhui Li}
\ead{18700757288@163.com}
\credit{Conceptualization of this study, Methodology, Software}
\fnmark[5]
\fntext[5]{Junhui Li is a student in  China Civil
Aviation Flight Academy of China University}
\cormark[2]
\cortext[2]{Wuyang Luan. School of Computer,China Civil Aviation Flight Academy,Guanghan City, Deyang Province, Sichuan,Guanghan City, Deyang Province, Sichuan,China.}

\affiliation[1]{organization={China Civil Aviation Flight Academy},
                city={Guanghan City, Deyang Province, Sichuan},
                postcode={618307}, 
                country={China}}

\begin{abstract}
Most existing 3D instance segmentation methods are derived from  3D semantic segmentation models. However, these indirect approaches suffer from certain limitations. They fail to fully leverage global and local semantic information for accurate prediction, which hampers the overall performance of the 3D instance segmentation framework. To address these issues, this paper presents PSGformer, a novel 3D instance segmentation network. PSGformer incorporates two key advancements to enhance the performance of 3D instance segmentation.
Firstly, we propose a Multi-Level Semantic Aggregation Module, which effectively captures scene features by employing foreground point filtering and multi-radius aggregation. This module enables the acquisition of more detailed semantic information from global and local perspectives.
Secondly, PSGformer introduces a Parallel Feature Fusion Transformer Module that independently processes super-point features and aggregated features using transformers. The model achieves a more comprehensive feature representation by the features which connect global and local features. 
We conducted extensive experiments on the ScanNetv2 dataset. Notably, PSGformer exceeds compared state-of-the-art methods by 2.2\% on ScanNetv2 hidden test set in terms of mAP. Our code and models will be publicly released.
\end{abstract}

\maketitle

\section{Introduction}
\label{}
3D Instance Segmentation \cite{tang2022bi,qayyum2023semi,zhao2022dsu,dayananda2023amcc,nakamura2021effective,cheng2021ptanet} is a pivotal task in the field of computer vision, with the primary goal of identifying and segmenting individual entities within 3D point cloud data. It is required to not only ascertain the category of each object, but also differentiate between distinct entities of the same class. This technology plays a significant role in a myriad of practical applications, such as autonomous driving  \cite{zhou2020joint}, robotic navigation \cite{xie2021unseen}, augmented reality/virtual reality  \cite{park2020deep}, and 3D modeling, to name a few. 

There are two predominant strategies in the current field of 3D instance segmentation: proposal-based \cite{yang2019learning,qi2017pointnet++,hou20193d} and grouping-based \cite{wang2019associatively,wang2018sgpn,elich20193d,lahoud20193d,liu2019masc,qi2019deep}. The proposal-based approach, as demonstrated in the work of researchers like Yang and Liu, is a top-down method that initially posits a series of region proposals (or bounding boxes), and subsequently predicts the shape of instances within these regions. This approach borrows from the successful experience of Mask-RCNN \cite{he2017mask} in 2D instance segmentation. However, it encounters difficulties when dealing with point cloud data. In 3D environments, bounding boxes have more degrees of freedom, making fitting more challenging \cite{yang2019learning}. Also, as point clouds only exist on a part of an object's surface, the geometric centers of objects are difficult to detect. Furthermore, low-quality region proposals can also negatively impact overall model performance.

Conversely, the grouping-based approach, represented by the work of Jiang et al \cite{jiang2020pointgroup}, Chen et al \cite{chen2021hierarchical}, Liang et al \cite{liang2021instance}, and Vu et al \cite{vu2022softgroup}, among others, is a bottom-up method. They learn the semantic labels and instance center offsets for each point, and then use these offset points and semantic predictions for instance aggregation. This approach has made significant progress in the 3D instance segmentation task in the past two years. Nevertheless, it also comes with some limitations: Grouping-based methods heavily rely on their semantic segmentation results, and erroneous predictions can suppress network performance. 

\begin{figure*}[pos=!h]
\centering
\begin{subfigure}{.3\textwidth}
  \centering
  \includegraphics[width=.8\linewidth]{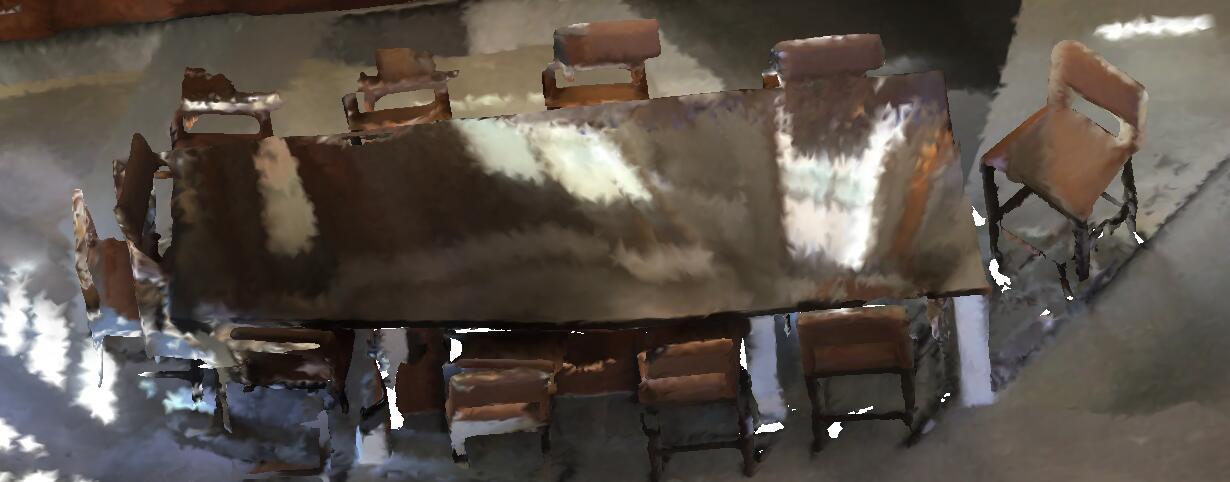}
  \caption{Input Point Cloud}
  \label{fig:sub1}
\end{subfigure}%
\begin{subfigure}{.3\textwidth}
  \centering
  \includegraphics[width=.8\linewidth]{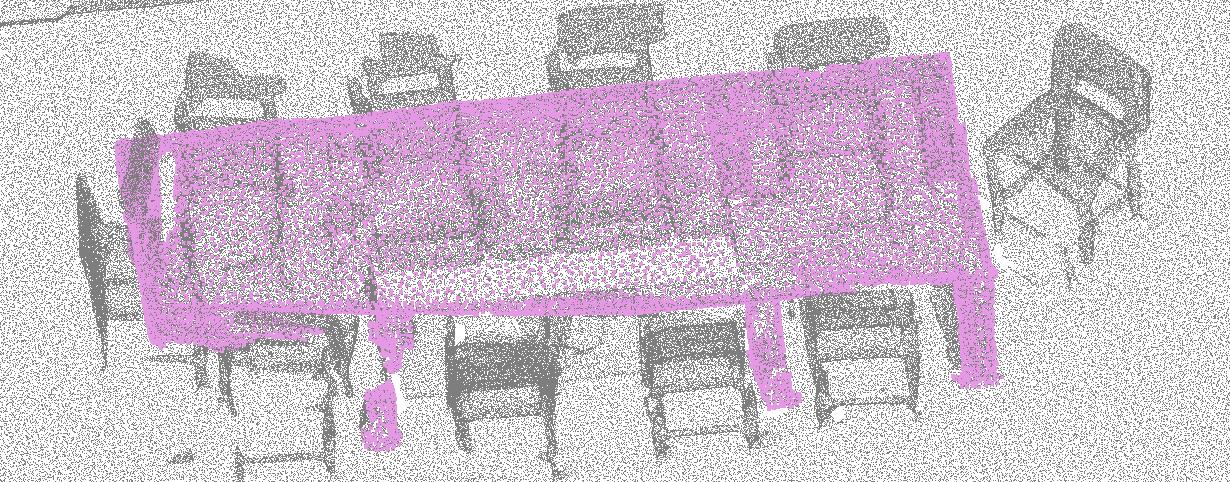}
  \caption{3D Instance}
  \label{fig:sub2}
\end{subfigure}
\begin{subfigure}{.3\textwidth}
  \centering
  \includegraphics[width=.8\linewidth]{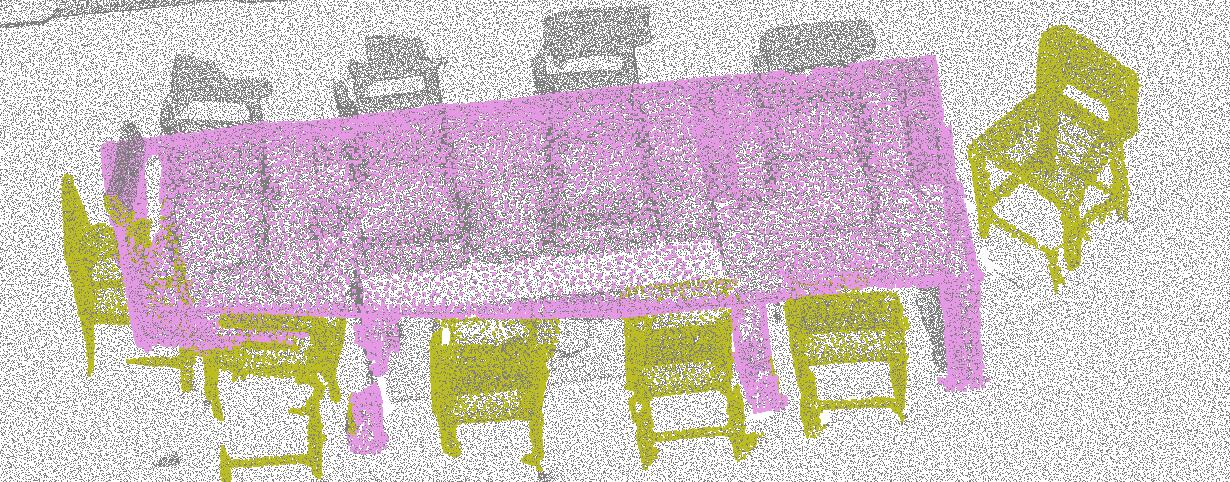}
  \caption{3D Semantic Instances}
  \label{fig:sub3}
\end{subfigure}
\newline
\begin{subfigure}{.3\textwidth}
  \centering
  \includegraphics[width=.8\linewidth]{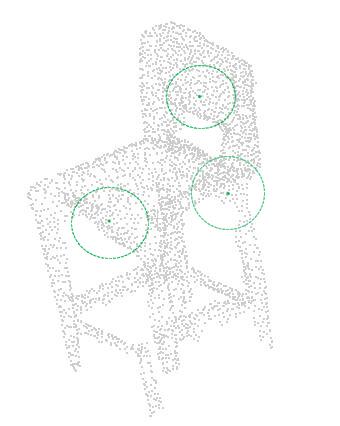}
  \caption{Grouping-based method}
  \label{fig:sub4}
\end{subfigure}%
\begin{subfigure}{.3\textwidth}
  \centering
  \includegraphics[width=.8\linewidth]{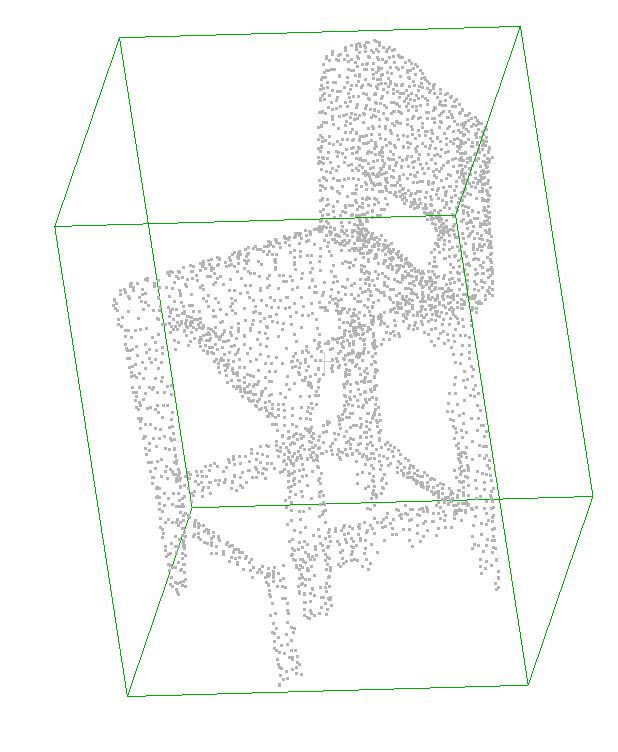}
  \caption{Proposal-based method}
  \label{fig:sub5}
\end{subfigure}
\begin{subfigure}{.3\textwidth}
  \centering
  \includegraphics[width=.8\linewidth]{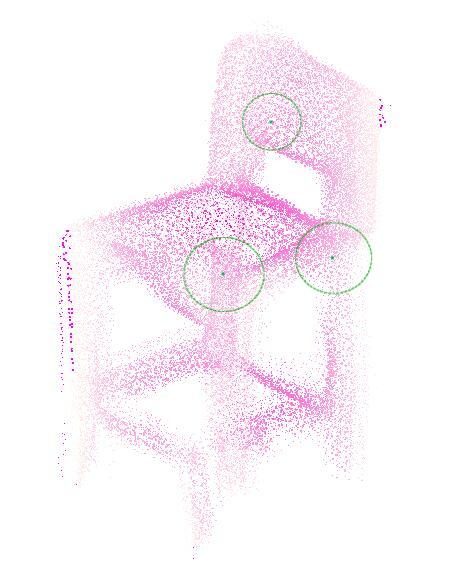}
  \caption{Ours}
  \label{fig:sub6}
\end{subfigure}
\caption{The introduction figure provides a comprehensive comparison of our method with conventional ones. (a) represents the input point cloud data. (b) shows an example of an individual instance segmentation approach, in which each object is separately recognized and labeled. (c) represents a 3D instance segmentation example where objects are recognized and classified in a three-dimensional space. (d) presents the results of a grouping-based method, which groups the points based on their spatial relationships and similarities. (e) illustrates a proposal-based approach where proposals for possible object locations are generated prior to recognition and classification. Lastly, (f) showcases our method, which integrates the benefits of both grouping and proposal-based methods and further enhances the performance through cross-attention mechanism. This visual representation helps to emphasize the effectiveness and superiority of our approach in handling point cloud data.}
\label{fig:Schematic drawing}
\end{figure*}

In response to these challenges, we propose a novel solution in this study that leverages the powerful representation of both local and global scene features. Specifically, we introduce a Multi-Level Semantic Aggregation Module, which effectively captures local scene features by employing foreground point filtering and multi-radius aggregation. In addition, we utilize a superpoint pooling method to obtain global, superpoint-level features. These two types of features are independent before processing and fusion, allowing the model to fully exploit their respective advantages and characteristics. Such a design alleviates the difficulties encountered by proposal-based methods in handling point cloud data and enhances object detection accuracy. Next, we combine these two independent features into a powerful feature representation through our Parallel Feature Fusion Transformer module, considering both global information and not neglecting local information.

To evaluate the performance of our proposed approach, we conducted extensive experiments on the challenging ScanNetV2 dataset. Our method achieves state-of-the-art on this dataset, surpassing the most potent method currently available. In summary, our work's main contributions can be distilled as follows:

• As for our proposed Multi-Level Semantic Aggregation Module, it is designed to capture and utilize semantic information more effectively at different levels. This module excels at multi-scale feature learning and can fully exploit the hierarchical nature of point cloud data, thus effectively capturing and integrating multi-level and multi-scale feature information. In this process, the module aggregates features within different radius ranges, grasping rich semantic information from local perspectives. Consequently, it enhances the representational capacity and generalizability of our model, especially regarding the extraction of local features, while mitigating the over-reliance on semantic segmentation results.

• We designed a parallel feature fusion Transformer architecture to better process and fuse information from different sources. The structure was designed with the importance and differences of local and global features in mind, and how to integrate the two features more effectively. By processing and fusing these features in parallel, our model is able to focus on both global and local information, avoiding possible information loss in multiple transitions. This parallel feature fusion mechanism allows the model to use and fuse information from different levels and scales more efficiently, thereby improving the model's predictive performance.

• Our method not only achieves state-of-the-art performance on the ScanNetV2 dataset, but also surpasses the previous best result by 2.2\% mAP, which further substantiates the effectiveness of our approach.

\section{Related Work}

\textbf{Proposal-based methods.} Many approaches \cite{yang2019learning, hou20193d, qi2017pointnet++} follow a two-step pipeline. Initially, they detect instance candidates, often as 3D bounding boxes, and then refine the instances within these bounding boxes. Methods such as 3D-BoNet  \cite{yang2019learning} and 3D-SIS \cite{hou20193d} employ this strategy. For instance, 3D-BoNet \cite{yang2019learning} utilizes PointNet++ \cite{qi2017pointnet++} to extract point cloud features, and sees the 3D bounding box generation task as an optimal assignment problem. HPGN \cite{shen2021exploring} proposes a novel pyramid graph network targeting features, which is closely connected behind the backbone network to explore multi-scale spatial structural features.  However, instead of relying on predefined anchor  boxes like 3D-SIS \cite{hou20193d}, 3D-BoNet \cite{yang2019learning} predicts bounding boxes \cite{shen2021competitive} from a global scene descriptor and optimizes an association loss based on bipartite matching. The key challenge for these methods is that the quality of the instance segmentation is significantly affected by the accuracy of bounding box predictions.

\textbf{Grouping-based methods.} This approach treats 3D instance segmentation as a process of feature learning followed by instance grouping. Methods like ASIS \cite{wang2019associatively}, SGPN \cite{wang2018sgpn}, and 3D-BEVIS \cite{elich20193d} employ contrastive learning, mapping points to a high-dimensional feature space where features of the same instance are close together, and far apart otherwise. Other techniques like MTML \cite{lahoud20193d} and MASC \cite{liu2019masc} have further improved performance by leveraging powerful feature backbones like sparse convolutional networks. The EMRN \cite{shen2021efficient} proposes a multi-resolution features dimension uniform module to fix dimensional features from images of varying resolutions.  Additionally, inspired by Hough voting approaches, methods such as VoteNet \cite{qi2019deep} proposed center-voting for 3D object detection, shifting from mapping points to a high-dimensional feature space to a method where points vote for their object center \cite{qiao2022novel}.  This facilitates the formation of geometrically coherent instance masks, resulting in more accurate segmentation outcomes. Nonetheless, such methods typically necessitate an additional intermediate aggregation stage, thereby increasing the time and computational complexity involved in both training and inference processes.

\textbf{Instance Segmentation with Transformer.} In the realm of 2D instance segmentation, the power of Transformers \cite{song2023dpctn} has been harnessed in several state-of-the-art works. For instance, DETR  \cite{zhu2020deformable} has demonstrated superior performance for various vision tasks \cite{li2022enhancing, shen2023pbsl}, owing to the Transformer's inherent capability to model long-range dependencies which is beneficial for handling complex scenes. More recently, MaskFormer \cite{li2022maskformer}, a simple and efficient Transformer-based method for instance segmentation, showed remarkable results in both 2D and 3D instance segmentation tasks.
However, the extension of Transformers to 3D instance segmentation isn't straightforward and presents unique challenges. 3D point cloud data, unlike 2D images, consist of non-grid data, making it hard for conventional Transformers \cite{shen2023git}, which were originally designed for grid-like data. Moreover, the volumetric nature and high-dimensionality of 3D data demand more computational resources, raising scalability and efficiency concerns. In light of these challenges, it becomes crucial to develop Transformer architectures that are tailored to the specifics of 3D data.

Drawing upon the immense potential of the Transformer architecture, we present a pioneering framework for 3D instance segmentation called PSGformer. Our primary objective is to adeptly capture and leverage diverse levels of semantic information, while also effectively processing and integrating various feature data. To accomplish this, we introduce a multi-layer semantic aggregation module that enables the extraction of local scene features through foreground point filtering and multi-radius aggregation. Additionally, we employ hyperpoint pooling to obtain comprehensive global hyperpoint level features. Furthermore, we have devised a parallel feature fusion Transformer structure. By concurrently processing and fusing these feature data, our model adeptly incorporates both global and local information, thereby circumventing any loss of information that may arise during multiple transitions. Consequently, our framework not only addresses the challenge of applying Transformers to 3D point cloud data but also capitalizes on the Transformers' inherent capacity to enhance the predictive performance of the model.

\section{Method}

\subsection{Overall}
\begin{figure*}[pos=!h]
\centering
\includegraphics[width=\textwidth]{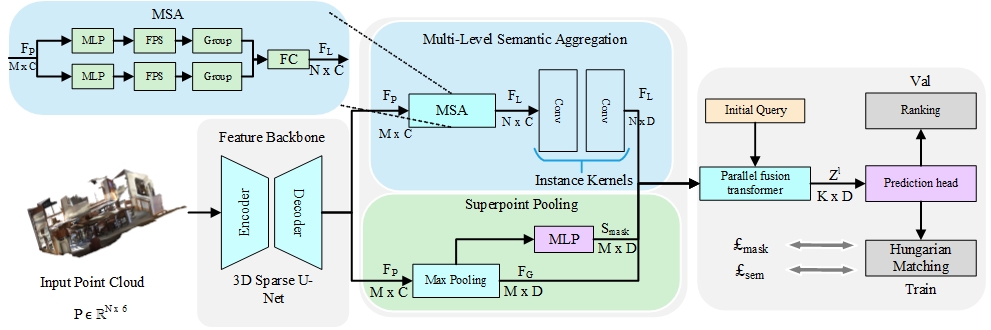}
\caption{We propose an end-to-end 3D instance segmentation network designed for the precise and efficient handling of 3D point cloud data. As shown in the figure, our network framework consists of the following key steps:(a) The point cloud data is first fed into a sparse 3D U-Net structure for per-point feature extraction. (b) The network then proceeds via two distinct paths to obtain features: one path generates local instance feature points through multi-level semantic aggregation, while the other obtains global superpoint features via superpoint max-pooling. (c) These two types of features are fused and fed into a Transformer decoder to generate advanced query vector features. (d) Within the prediction head, these query vectors are utilized to generate the predicted instance masks. (e) During the training stage, the predicted masks are binary matched with the ground truth masks to calculate the loss. During the prediction stage, the predicted masks are sorted to obtain the final instance segmentation outcomes.}
\label{fig:network}
\end{figure*}

As shown in Figure \ref{fig:network},  we firstly employ a sparse 3D U-net to extract bottom-up point-wise features. Subsequently, potential point-wise features are grouped into superpoints via a simple superpoint pooling layer, serving as global features. Following this, we introduce a multi-level semantic aggregation module that employs foreground point selection and dual-radius clustering to further refine the cloud points, thereby obtaining local features. The processed local and global features are then fed into our newly designed Parallel Feature Fusion Transformer structure. This novel structure enhances the Transformer’s ability to handle and integrate diverse feature information, whilst capturing instance information through cross-attention mechanism. Finally, through bipartite matching-based on superpoint masks, PSGformer can implement end-to-end training.

\subsection{Feature Backbone}
In the encoder part of our network architecture, a Sparse 3D U-Net  \cite{graham20183d} model is employed to extract the features of the input 3D point cloud on a point-by-point basis. The Sparse 3D U-Net consists of sequential encoding and decoding blocks, with skip connections between layers to capture multi-scale information. To achieve sparsity, we incorporate a sparsification operation during the training process, which allows the network to process only significant parts of the point cloud data, thus significantly improving computational efficiency. Through this Sparse 3D U-Net, we can effectively extract rich feature representations $F_{P} \in R^{N \times C}$ from the input point cloud $P \in R^{N \times 6}$, which serves as a foundation for subsequent multi-level semantic aggregation.

\subsection{Multi-level  Semantic Aggregation}

Our Multi-level Semantic Aggregation Module operates as an independent component, markedly reducing computational overhead and alleviating redundancy in data processing. The module initially applies an efficient foreground point filtration method. This technique filters points based on the probability of each point being predicted as foreground, represented as $m(i)(0) \in [0,1]$. The Farthest Point Sampling (FPS) strategy is implemented, selecting candidate points from the point set based on the predicted foreground mask, $1-m(i)(0)> \beta$. This process not only minimizes redundancy but also optimizes computational load. Herein, we introduce an iterative sampling technique, sampling candidates from a set of points that are neither background nor chosen by previously sampled candidates. This can be formally represented as follows:

\begin{equation}
P'=\{p(i) \in P | \min_{k=0..K'}(1-m(i)(k))> \beta\} ,
\end{equation}
where $K'$ denotes the number of candidates already selected, $P'$ is the filtered point cloud and $\beta$ is the hyper-parameter threshold. This strategy essentially optimizes the sampling of foreground points, enabling the sampling process to cover all instances maximally, irrespective of their sizes.

Following the filtration and sampling of foreground points, we employ the Farthest Point Sampling (FPS) and Sphere Query process, extracting rich semantic information for each local keypoint through the following formula:

\begin{equation}
Q = \{ q(i) \in Q | \text{distance}(q(i), p(j)) < r , \text{for any} , p(j) \in P\} ,
\end{equation}
here, $Q$ is the set of keypoints, distance(·) refers to the Euclidean distance, and $r$ is the radius of the sphere query.

Simultaneously, a superpoint max-pooling operation is executed in parallel, generating global superpoint features. These parallel operations yield a compact yet highly informative feature representation.

Upon the extraction of keypoint information, the module then stacks multiple convolution blocks, generating high-dimensional features and further enriching the representation of the features.

Suppose our input voxel-level point cloud feature is $F_{P} \in R^{M \times 6}$, where $N$ is the number of points in the point cloud. The two radii for spherical queries are $r_1$ and $r_2$, and the two orientations are $\theta_1$ and $\theta_2$. The semantic scores of the foreground points are calculated by a linear layer, with a threshold set to $t$. Only points with scores greater than $\tau$ will be selected as foreground points. The extraction of key points is carried out through the furthest point sampling and spherical query, with a sampling radius of $r_s$ and a query radius of $r_q$. The final linear layer integrates these extracted features together, outputting features as $F_{L} \in R^{N \times C}$, where $M$ is the number of key points and $C$ is the dimension of the features.

\subsection{Superpoint Pooling}
In our model, we incorporate a SuperPoint Pooling module to more effectively handle and integrate the input features. Specifically, this module takes the features $F_{P}$, extracted by the sparse 3D U-Net, as inputs. We then adopt a precomputed superpoints strategy and directly feed these point-wise features into the superpoint pooling layer. The superpoint pooling layer acquires the superpoint features by performing an average pooling operation over the point-wise features within each superpoint, allowing us to derive M superpoints from the input point cloud.

Notably, through this approach, the superpoint pooling layer reliably downsamples the input point cloud to hundreds of superpoints, thereby significantly reducing the computational overhead of subsequent processing while optimizing the representational capacity of the entire network. On the other hand, by processing the real mask $S_{mask}$ with a Multilayer Perceptron (MLP) and obtaining the global superpoint features $F_{G}$ through the superpoint pooling layer, we further enhance the performance and accuracy of our model. Overall, this processing and integration approach enables our model to effectively leverage global information and achieve higher accuracy.

\subsection{Parallel fusion transformer}

\begin{figure}[t]
\centering
\includegraphics[width=0.5\textwidth]{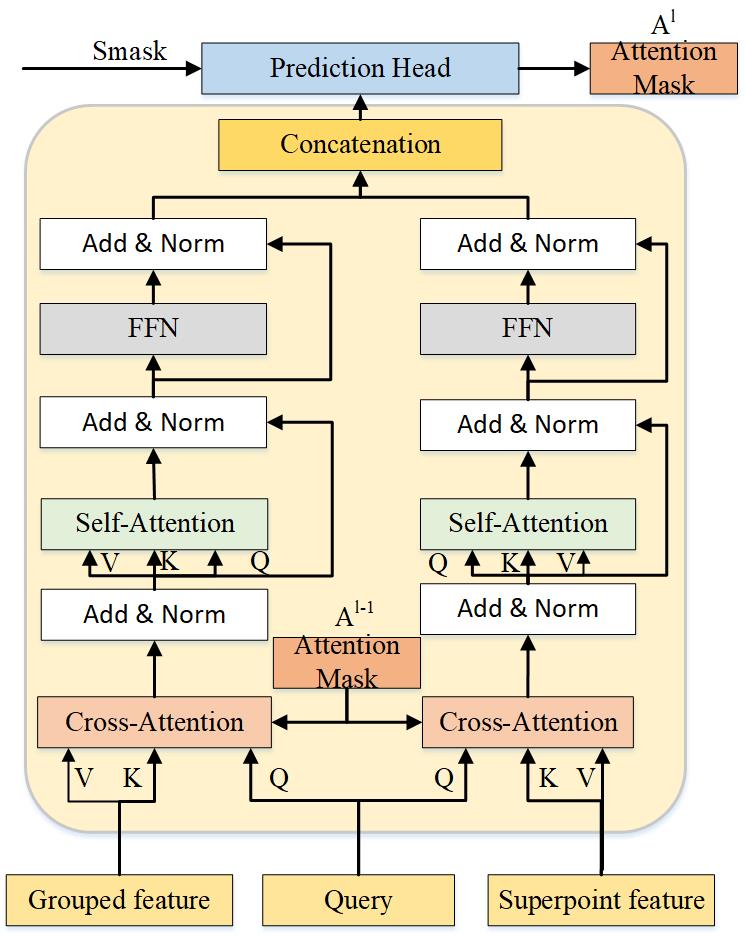}
\caption{The parallel fusion Transformer is the core of our decoder architecture, composed of a series of Transformer decoder layers that iteratively learn and update learnable query vectors through cross-attention mechanisms with global superpoint features and local features. }
\label{fig:transformer}
\end{figure}

As shown in Figure. \ref{fig:transformer}. , Our network architecture's feature decoder primarily consists of three components: a Transformer module, a prediction head, and a final matching and ranking procedure. The decoder receives two main inputs: the keypoint features $F_L \in R^{N \times C}$ extracted by the Multi-Level Semantic Aggregation (MSA) module, and the global superpoint features $F_{G} \in R^{M \times D}$ obtained through SuperPoint pooling.

These features are initially processed in parallel by the Transformer structure, where they are learned via query vectors. We assume there are $K$ learnable query vectors. The features of the query vectors from each Transformer decoder layer are predefined as $Z_l \in R^{K \times D}$, where $D$ is the embedding dimension, and $l$ is the layer index.

After processing in the Transformer, the output results are directed into the prediction head, compared with the true masks$S_{mask} \in R^{M \times D}$, and produce the final predicted masks. In the training stage, we implement a bipartite matching procedure using the Hungarian algorithm. During inference, we use a top-k ranking procedure. This way, our feature decoder effectively employs both local and global features to perform efficient segmentation of 3D point cloud instances.

The parallel fusion Transformer decoder, it is constructed with some unique innovations. The global superpoint feature $F_G \in R^{M \times D}$ and the local aggregated feature $F_L \in R^{N \times C}$ serve as two parallel input channels, respectively interacting with the learnable query vectors $A^{l-1} \in R^{K \times D}$ via the superpoint cross-attention operation. Subsequently, the self-attention operation and feed-forward neural network (FFN) further enrich the information of the query vectors, generating a new set of query vectors $A^l$. Ultimately, the query vectors generated from both parallel channels are output through a fully connected layer.

The superpoint cross-attention mechanism can be expressed by the following formula:
\begin{center}
\begin{equation}
\hat{Z}^l = \text{softmax}(QK^T/\sqrt{D} + A^{l-1})V ,
\end{equation}
\end{center}
here, $\hat{Z}^l \in R^{K \times D}$ is the output of the superpoint cross-attention. $Q = \psi_Q(Z^{l-1})$ is the linear projection of input query vectors $Z^{l-1}$ and $K$ and $V$ are different linear projections, $\psi_K$ and $\psi_V$, of the superpoint feature $F_G'$. $A^{l-1} \in R^{K \times M}$ is the superpoint attention mask.

Regarding the superpoint attention mask, it utilizes the superpoint mask $M^{l-1}$ generated from the former prediction head and filters it through a threshold $\tau$. This can be expressed as:

\begin{equation}
A^{l-1}(i, j) = \begin{cases} 0, & \text{if } M^{l-1}(i, j) \geq \tau \\ -\infty, & \text{otherwise} \end{cases} ,
\end{equation}

We empirically set the threshold $\tau$ to 0.5. In the stacking of the Transformer decoder layers, the superpoint attention mask $A^{l-1}$ adaptively constrains the cross-attention within the foreground instance. Specifically, $A^{l-1}(i, j)$ indicates the $i$-th query vector paying attention to the $j$-th superpoint, when and only when $M^{l-1}(i, j)$ exceeds the threshold $\tau$.Such design enables the attention mask $A^{l-1}$ to adaptively constrain the cross-attention within the foreground instance in each iteration.

\subsection{Prediction Head.}
With the obtained query vectors $Z^{l}$ from the instance branch, we utilize two independent MLPs. One predicts the classification ${p_i} \in R^{N_{class}+1}, i = 1, 2, \ldots, K$ for each query vector, which helps evaluate the quality of the proposals via an IoU-aware scoring mechanism ${s_i} \in [0, 1], i = 1, 2, \ldots, K$. Specifically, we append a 'no instance' probability in addition to $N_{class}$ categories to arrange the ground truths to the proposals by bipartite matching, treating the remaining proposals as negative predictions. Further, due to the potential discrepancy in the ranking of proposal quality, we propose a scoring branch which estimates the IoU between the predicted superpoint masks and ground truth masks. This allows us to adjust for any potential misalignment in the proposal quality ranking. Lastly, by using the mask-aware features $S_{mask} \in R^{M \times D}$ from the mask branch, we directly multiply it with the query vectors $Z^{l}$ and apply a sigmoid function to generate superpoint masks prediction $M^{l} \in [0, 1]^{K \times M}$.

\subsection{Loss function}

optimization is primarily reliant on four key losses: the classification loss, score loss, and two mask losses - the Binary Cross-Entropy (BCE) loss and the Dice loss. Each of these losses corresponds to a specific functionality in the model, working together to enhance the model's predictive accuracy.

Firstly, the classification loss mainly optimizes the model's classification capabilities. Using the Cross-Entropy loss, we compute the divergence between the predicted classification labels and the groundtruth labels. By minimizing this divergence, we enhance the model's classification accuracy. The formula for the Cross-Entropy loss as 

\begin{equation}
L_{cls} = -\frac{1}{N}\sum_{i=1}^{N} y_i \log(p_i) + (1-y_i) \log(1-p_i) ,
\end{equation}
where $y_i$ is the groundtruth label, and $p_i$ is the predicted probability.

Next, the score loss measures the discrepancy between the model's predicted scores and the true scores calculated by Intersection over Union (IoU). We use the Mean Squared Error (MSE) loss to compute this discrepancy and improve the model's score prediction accuracy by minimizing it. The formula for the MSE loss is 

\begin{equation}
L_{score} = \frac{1}{N}\sum_{i=1}^{N} (s_i - \hat{s}_i)^2 ,
\end{equation}
where $s_i$ is the groundtruth score, and $\hat{s}_i$ is the predicted score.

For mask losses, we use both the Binary Cross-Entropy loss \cite{ruby2020binary} and the Dice loss \cite{li2019dice}. The Binary Cross-Entropy loss \cite{ruby2020binary} measures the divergence between the predicted masks and the groundtruth masks, while the Dice loss \cite{li2019dice} gauges the similarity between the predicted masks and the groundtruth masks. By minimizing these two losses, we enhance the model's mask prediction accuracy. The formula for the BCE loss as 

\begin{equation}
L_{BCE} = -\frac{1}{N}\sum_{i=1}^{N} y_i \log(p_i) + (1-y_i) \log(1-p_i) ,
\end{equation}
where $y_i$ is the groundtruth mask value, and $p_i$ is the predicted mask value. The formula for the Dice loss  \cite{milletari2016v} as 
\begin{equation}
L_{Dice} = 1 - \frac{2 \cdot |P \cap G|}{|P| + |G|} ,
\end{equation}
where $P$ and $G$ represent the predicted masks and groundtruth masks, respectively.

In conclusion, our overall optimization objective function is a weighted sum of these four losses, with the formula being 
\begin{equation}
L_{total} = w_{cls} L_{cls} + w_{score} L_{score} + w_{BCE} L_{BCE} + w_{Dice} L_{Dice} ,
\end{equation}
where $w_{cls}$, $w_{score}$, $w_{BCE}$, and $w_{Dice}$ are the respective weights for each loss. Specifically, the weights are set as 0.5 for classification, 0.5 for scoring, and 1 for masking.

During the inference stage, given an input point cloud, our model directly predicts K instances, each with a classification ${p_i}$, an IoU-aware score ${s_i}$, and corresponding superpoint masks. Moreover, we obtain a mask score ${ms_i \in [0, 1]}^K$ by averaging superpoint probabilities that exceed 0.5 in each superpoint mask. Lastly, to conduct sorting, we calculate a final score $\tilde{s}_i = \sqrt[3]{p_i \cdot s_i \cdot ms_i}$. Our model does not require non-maximum suppression during post-processing, thereby ensuring swift inference speed.
\section{Experiments}

\textbf{Datasets.} ScanNetv2 \cite{dai2017scannet} is a large-scale 3D indoor scenes dataset that contains 1513 scanned scenes, each consisting of a series of RGB-D images along with corresponding sensor data. The dataset provides instance-level semantic annotations for 3D point clouds through a semi-automated annotation process, covering 40 common indoor object categories. ScanNet v2 offers a rich and detailed data foundation for 3D vision tasks.The Stanford Large-Scale 3D Indoor Spaces (S3DIS) \cite{armeni20163d} dataset encompasses 3D scans of six large commercial buildings, totalling 272 distinct indoor spaces. The data for each space includes RGB-D images and 3D geometric data. All 3D point cloud data are labelled with 13 common indoor element categories, such as walls, ceilings, floors, windows, etc. The S3DIS dataset provides detailed and varied data support for 3D vision tasks in indoor scenarios.

\textbf{Evaluation index.} We adopt the Task-mean Average Precision (mAP) as the primary evaluation metric for instance segmentation. This method averages the scores over a range of IoU thresholds, starting from 50\% and ending at 95\%, with an interval of 5\%. More specifically, AP50 and AP25 represent the scores corresponding to the IoU thresholds of 50\% and 25\%, respectively. We report these metrics, namely mAP, AP50, and AP25, on the ScanNetv2 dataset. 

\subsection{Benchmark Results}
\textbf{ScanNetv2}. 
As shown in \ref{table:scannetv2 hadden test}, we compared PSGFormer with existing high-performance methods on the hidden test set. PSGFormer achieves an mAP score of 55.4\%. For the specific 18 categories, our model achieves the highest mAP scores in several of them. Particularly in the counter category, which past methods have found challenging, PSGFormer surpasses the previous model by more than 2.2\%. We also evaluated PSGFormer on the ScanNetv2 validation set, as demonstrated in \ref{table:scannetv2 val set}. PSGFormer performs well among all high-performance methods. Compared to the next-best results, our method shows improvements in terms of mAP, AP50, and AP25.
This can largely be attributed to our two key innovations: the multi-layer semantic aggregation module and the parallel fusion Transformer structure.
Firstly, aiming to capture precise characteristics in both global and local scene features, we propose a multi-layer semantic aggregation module. This module initially obtains features rich in local information through foreground point filtering and multi-radius aggregation. Subsequently, we employ superpoint pooling to grasp global features from a more macro perspective. These two kinds of features remain independent before being processed and fused, allowing the model to extract and utilize the advantages and characteristics of each, achieving a more comprehensive understanding of the scene.

Secondly, in order to fully tap the potential of these two types of features and retain their individual advantages, we introduce a parallel fusion Transformer structure. This structure allows global and local features to be fused in parallel while maintaining their uniqueness, avoiding potential information loss in multiple transformations. The parallel fusion Transformer structure can effectively integrate information at various scales and levels. In this way, the model can more comprehensively understand and depict the scene, thereby performing better in various categories, including the challenging counter category.

Particularly in the task of handling the counter category, PSGFormer surpasses the previous model by more than 2.2\%. This significant improvement demonstrates that our multi-layer semantic aggregation module and parallel fusion Transformer structure can effectively alleviate the difficulties of traditional models in dealing with complex categories, thus greatly enhancing the performance of the model.
\begin{table*}[pos=!h]
\centering
\resizebox{1\textwidth}{!}{%
\begin{tabular}{c|c|c c c c c c c c c c c c c c c c c c}
\hline
Method & mAP  & bath  &  bed  &  bkshf & cabinet  & chair & counter & curtain & desk & door & other & picture & fridge & s. cur. 
 & sink & sofa & table & toilet & wind.\\
\hline
3D-BoNet  \cite{yang2019learning} & 25.3  &51.9 & 32.4 & 25.1& 13.7 &34.5& 3.1& 41.9& 6.9& 16.2 &13.1& 5.2& 20.2 &33.8& 14.7& 30.1& 30.3 &65.1 &17.8 \\
MTML  \cite{lahoud20193d} & 28.2 &57.7& 38.0& 18.2& 10.7 &43.0& 0.1 &42.2& 5.7& 17.9 &16.2& 7.0 &22.9 &51.1 &16.1 &49.1 &31.3& 65.0 &16.2 \\
3D-MPA  \cite{engelmann20203d} &35.5 &45.7 &48.4 &29.9 &27.7& 59.1 &4.7& 33.2& 21.2& 21.7 &27.8 &19.3& 41.3& 41.0 &19.5& 57.4&35.2 &84.9 &21.3 \\
PointGroup  \cite{jiang2020pointgroup} & 40.7 &63.9 &49.6 &41.5 &24.3& 64.5& 2.1& 57.0& 11.4 &21.1& 35.9 &21.7& 42.8& 66.0& 25.6& 56.2 &34.1 &86.0 &29.1\\
OccuSeg  \cite{han2020occuseg} & 48.6 & 80.2& 53.6& 42.8& 36.9 &70.2& 20.5& 33.1& 30.1& 37.9 &47.4& 32.7 &43.7& 86.2 &48.5& 60.1& 39.4& 84.6 &27.3\\
DyCo3D  \cite{he2021dyco3d} &39.5& 64.2& 51.8 &44.7 &25.9& 66.6& 5.0 &25.1& 16.6 &23.1& 36.2 &23.2& 33.1 &53.5 &22.9 &58.7 &43.8 &85.0& 31.7\\
PE  \cite{zhang2021point} & 39.6 &66.7& 46.7& 44.6 &24.3 &62.4 &2.2 &57.7& 10.6 &21.9 &34.0& 23.9& 48.7& 47.5& 22.5& 54.1& 35.0 &81.8& 27.3\\
HAIS  \cite{chen2021hierarchical} &45.7 &70.4& 56.1 &45.7 &36.3& 67.3& 4.6& 54.7 &19.4 &30.8 &42.6 &28.8& 45.4 &71.1 &26.2& 56.3& 43.4& 88.9 &34.4\\
SSTNet  \cite{liang2021instance}&50.6& 73.8& 54.9& 49.7& 31.6& 69.3 &17.8 &37.7 &19.8& 33.0& 46.3 &57.6 &51.5 &\textbf{85.7}& 49.4 &63.7 &45.7& 94.3& 29.0\\
SoftGroup  \cite{vu2022softgroup} & 50.4 &66.7& 57.9& 37.2& 38.1 &69.4& 7.2& \textbf{67.7}& 30.3& 38.7& 53.1& 31.9& \textbf{58.2}& 75.4& 31.8 &64.3 &49.2 &90.7 &\textbf{38.8}\\
RPGN  \cite{dong2022learning} & 42.8 &63.0& 50.8& 36.7& 24.9& 65.8& 1.6& 67.3& 13.1 &23.4& 38.3& 27.0& 43.4& 74.8 &27.4 &60.9& 40.6& 84.2 &26.7 \\
PointInst3D  \cite{he2022pointinst3d} & 43.8 &\textbf{81.5}& 50.7& 33.8& 35.5& 70.3& 8.9 &39.0 &20.8& 31.3 &37.3 &28.8& 40.1 &66.6 &24.2 &55.3& 44.2 &91.3 &29.3 \\
DKNet  \cite{wu20223d} & 53.2 &\textbf{81.5} &62.4& 51.7 &37.7& \textbf{74.9}& 10.7& 50.9& 30.4& 43.7& 47.5 &58.1& 53.9& 77.5& 33.9& 64.0 &50.6& 90.1 &38.5\\

\hline
PSGformer & \textbf{55.4}\textcolor{red}{($\uparrow$2.2)} & 74.1 & \textbf{68.8} & \textbf{54.8}&\textbf{43.6}&73.3&\textbf{20.9}&53.8&\textbf{32.1}&\textbf{50.1}&\textbf{50.9}&\textbf{59.5}&40.7&66.7&\textbf{51.9}&\textbf{70.4}&\textbf{52.7}&\textbf{94.6}&38.2\\
\hline
\end{tabular}
}
\caption{The 3D instance segmentation outcomes on the ScanNetV2 hidden test set, as measured by mAP scores, are presented. The most outstanding results are highlighted in bold, while the runner-up results are underlined. Our newly introduced approach takes the lead in achieving the highest mAP, surpassing the previous benchmark.}
\label{table:scannetv2 hadden test}
\end{table*}

\begin{table}[t]
\centering
\begin{tabular}{c|c|c c}
\hline
Method  & mAP  &   $AP_{\text{50}}$  &   $AP_{\text{25}}$ \\
\hline
SGPN  \cite{wang2018sgpn} & 19.3 & 37.8 & 53.4 \\
PointGroup  \cite{jiang2020pointgroup} & 34.8 & 51.7 & 71.3\\
HAIS  \cite{chen2021hierarchical} & 43.5 & 64.4 & 75.6\\
DyCo3D  \cite{he2021dyco3d} & 40.6 & 61.0 & 72.9\\
SSTNet  \cite{liang2021instance} & 49.4 & 64.3 & 74.0\\
SoftGroup  \cite{vu2022softgroup} & 46.0 & 67.6 & 78.9\\
PE  \cite{zhang2021point}  & 39.6 & 64.5 & 77.6 \\
Di\&Co3D  \cite{zhao2022divide} & 47.7 & 67.2 & 77.2\\
DKNet  \cite{wu20223d} & 50.8 & 66.7 & 76.9 \\
\hline
PSGformer & \textbf{57.7}\textcolor{red}{($\uparrow$6.9)} & \textbf{74.9}\textcolor{red}{($\uparrow$8.2)} & \textbf{83.4}\textcolor{red}{($\uparrow$6.5)}\\
\hline
\end{tabular}
\caption{3D instance segmentation results on ScanNetV2 validation set.}
\label{table:scannetv2 val set}
\end{table}

\textbf{S3DIS.}
We evaluated PSGformer on the S3DIS dataset using Area 5 and six-fold cross-validation, respectively. As evidenced in \ref{table:s3dis val set}, PSGformer sets a new state-of-the-art benchmark concerning the AP50 metric. In line with the protocols employed in previous methodologies, we additionally reported mPrec and mRec. Our approach also achieves competitive results concerning the mPrec/mRec metrics. The results on the S3DIS dataset corroborate the generalization capacity of PSGformer.

\begin{table}[t]
\centering
\begin{tabular}{c|c c c}
\hline
Method  & AP50  &  mPrec  &  mRec \\
\hline
PointGroup  \cite{jiang2020pointgroup} & 57.8 & 55.3 & 42.4\\
DyCo3D  \cite{he2021dyco3d} & - & 64.3 & 64.2\\
SSTNet  \cite{liang2021instance} & 59.3 & 65.5 & 64.2\\
HAIS  \cite{chen2021hierarchical} & - & 71.1 & 65.0\\
SoftGroup  \cite{vu2022softgroup} & 66.1 & \textbf{73.6} & 66.6\\
DKNet  \cite{wu20223d} & - & 70.8 & 65.3 \\
\hline
PSGformer & \textbf{66.7}\textcolor{red}{($\uparrow$0.6)} & 71.9 & \textbf{67.3}\\
\hline
3D-BoNet \dag  \cite{yang2019learning} & - & 65.6 & 47.7\\
PointGroup \dag  \cite{jiang2020pointgroup} & 64.0 & 69.6 & 69.2\\
SSTNet \dag  \cite{liang2021instance} & 67.8 & 73.5 & \textbf{73.4}\\
HAIS \dag  \cite{chen2021hierarchical} & - & 73.2 & 69.4\\
RPGN \dag  \cite{dong2022learning} &  - &\textbf{84.5} & 70.5\\
SoftGroup \dag  \cite{vu2022softgroup} & 68.9 & 75.3 & 69.8\\
DKNet \dag  \cite{wu20223d} & - & 75.3 & 71.1 \\
\hline
PSGformer \dag & \textbf{69.3}\textcolor{red}{($\uparrow$0.4)} & 74.9 & 71.3\\
\hline
\end{tabular}
\caption{3D instance segmentation results on the S3DIS validation set. Methods without a \dag mark are evaluated on Area 5; methods with a \dag mark are evaluated using six-fold cross-validation.}
\label{table:s3dis val set}
\end{table}

\begin{figure}[pos=!h]
\centering
\includegraphics[width=0.5\textwidth]{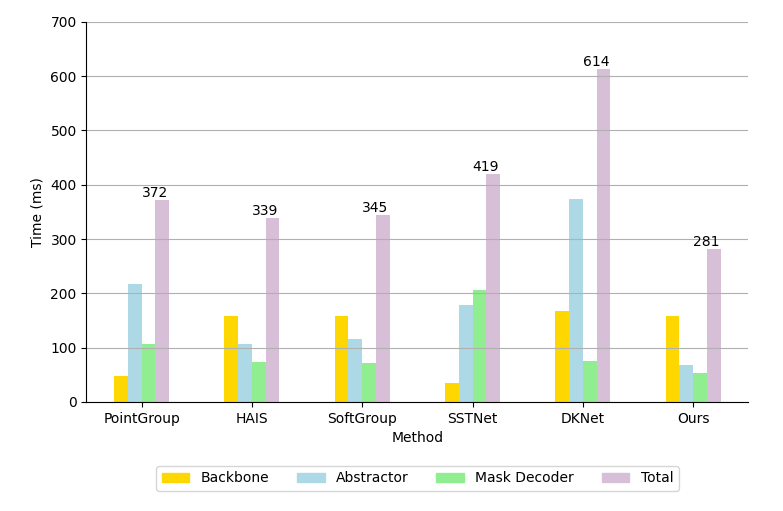}
\caption{Components and total runtimes (in ms) of five previous methods and PSGformer on ScanNetV2 validation set.}
\label{fig:runtimes}
\end{figure}

\textbf{Runtime Analysis}. 
Fig \ref{fig:runtimes}  present a comparison of the component and total runtime of our method and 5 other state-of-the-art methods for 3D instance segmentation, respectively, performed on a single 3080Ti GPU. These methods typically can be divided into three main components: the backbone, the instance abstractor, and the mask decoder. Our method proves to be the most efficient, with a total runtime of only 281ms and component runtimes of 159ms, 68ms, and 54ms for the backbone, instance abstractor, and mask decoder, respectively. In comparison to the instance abstractors used in PointGroup, DyCo3D, and SoftGroup, which rely on traditional clustering, our instance abstractor leverages our multi-layer semantic aggregation module, considerably accelerating aggregation, and hence reducing runtime. Additionally, our decoder is implemented using a transformer architecture. These findings underscore the efficiency and effectiveness of our proposed method.

\subsection{Ablation Study}

\textbf{The Architecture of Transformer.} 
Table \ref{table:ablation study}  explored the influence of three significant factors on the model's performance through ablation studies: the use of local features, global features, and the iterative use of the Transformer structure. Our experimental results indicate that when only local features are employed, the model's performance is noticeably inferior to the scenario where only global features are used. This affirms the vital role of global features in capturing overall point cloud information and structure.

However, we found that when local and global features are combined, the performance of the model significantly improves. This demonstrates that the combination of local and global features can capture richer and more complex features of point cloud data at both the global and local levels, which is highly valuable in the task of instance segmentation.

Furthermore, we observed that by iteratively using the Transformer structure, the performance of the model can be further improved. This is primarily attributable to the Transformer's self-attention mechanism, which can reveal deeper levels of feature interactions and dependencies, enabling the model to better understand and represent 3D point cloud data.

A crucial finding is that when we combine these three strategies: local features, global features, and iterative use of the Transformer structure, the model's performance is optimized. This result not only validates the effectiveness of our proposed method but also indicates that this approach, which combines global and local features and utilizes the Transformer structure, possesses potent representational capabilities. Lastly, to achieve optimal performance, we set the number of iterations for the Transformer structure to six in our experiments.segmentation.
\begin{table*}[pos=!h]
\centering
\begin{tabular}{c c c c |c c c }
\hline
No. & local feature& global feature & transformer cycle layers & mAP &  $AP_{\text{50}}$ &  $AP_{\text{25}}$  \\
\hline
 1& \checkmark & \ding{55}& \ding{55}& 43.5 & 54.8 & 60.8 \\
 2& \ding{55}& \checkmark &\ding{55} &50.9 & 65.7 & 76.1\\
 3& \checkmark &  \checkmark & \ding{55}& 54.5 & 71.0 & 78.7\\
 4&  \checkmark & \ding{55}&\checkmark & 57.2 & 74.8 & 82.9 \\
 5& \ding{55}& \checkmark & \checkmark&56.3 & 73.7 & 82.5\\
 6& \checkmark & \checkmark & \checkmark&\textbf{57.7} & \textbf{74.9} & \textbf{83.4}\\
\hline
\end{tabular}
\caption{The performance results of different choices of local features or global features.}
\label{table:ablation study}
\end{table*}

\textbf{Number of Queries and Layers.} 
The analysis of the table \ref{table:query and FPS sampling} shows that the model achieves the best mAP and AP50 performance with 400 query vectors and 1024 farthest distance samples.  Increasing the farthest distance samples or reducing the number of query vectors both lead to a decrease in performance.  Interestingly, the model with 400 query vectors and 512 farthest distance samples achieved the highest AP25, indicating a lower number of farthest distance samples may be beneficial for less strict precision requirements.  Therefore, careful tuning of these parameters is crucial, as they can significantly impact the model's performance depending on the task's precision requirements.

\begin{table}[pos=!h]
\centering
\begin{tabular}{c c |c c c }
\hline
Query & sampling & mAP &  $AP_{\text{50}}$ &  $AP_{\text{25}}$  \\
\hline
  400 & 2048 & 57.3 & 74.0 & 81.5 \\
  400 & 1024 & \textbf{57.7} & \textbf{74.9} & 83.4\\
  400 & 512  & 57.6 & 73.1 & \textbf{84.5}\\
\hline
  350 & 1024 & 56.9 & 72.9 & 82.5\\
  300 & 1024 & 57.1 & 73.4 & 83.7\\
  300 & 2048 & 56.3 & 71.5 & 81.0\\
\hline
\end{tabular}
\caption{The performance results of different choices of query vectors and number of FPS sampling.}
\label{table:query and FPS sampling}
\end{table}

\textbf{The Selection of Mask Loss.}
Table \ref{table:mask loss} provides an analysis of the individual contributions of different components of the mask loss function. Our findings indicate that relying solely on either binary cross-entropy loss or focal loss results in significant performance degradation. Evidently, the inclusion of Dice loss is crucial for optimizing mask loss. Furthermore, enriching Dice loss with either binary cross-entropy loss or focal loss enhances overall performance. Of all combinations tested, the pairing of Dice loss and binary cross-entropy loss proves to be the most effective.

\begin{table}[pos=!h]
\centering
\begin{tabular}{c c c |c c c}
\hline
Dice & Focal  &  BCE  &  mAP &  $AP_{\text{50}}$ &  $AP_{\text{25}}$  \\
\hline
 \ding{55} & \checkmark &\ding{55}   & 38.0 & 55.4 & 71.3 \\
 \ding{55} &\ding{55}   & \checkmark & 53.2 & 72.2 & 81.0\\
\checkmark &\ding{55}   & \ding{55} & 54.9 & 72.8 & 81.5\\
\checkmark & \checkmark &\ding{55}  &57.4 & 74.4 & 83.9\\
\checkmark &\ding{55}   & \checkmark & \textbf{57.7} & \textbf{74.9} & \textbf{83.4}\\
\hline
\end{tabular}
\caption{Ablation study on the selection of mask loss.}
\label{table:mask loss}
\end{table}

\subsection{Visualizations}
\begin{figure*}[pos=!h]
\centering
\begin{subfigure}{0.18\textwidth}
  \centering
  \includegraphics[width=\linewidth]{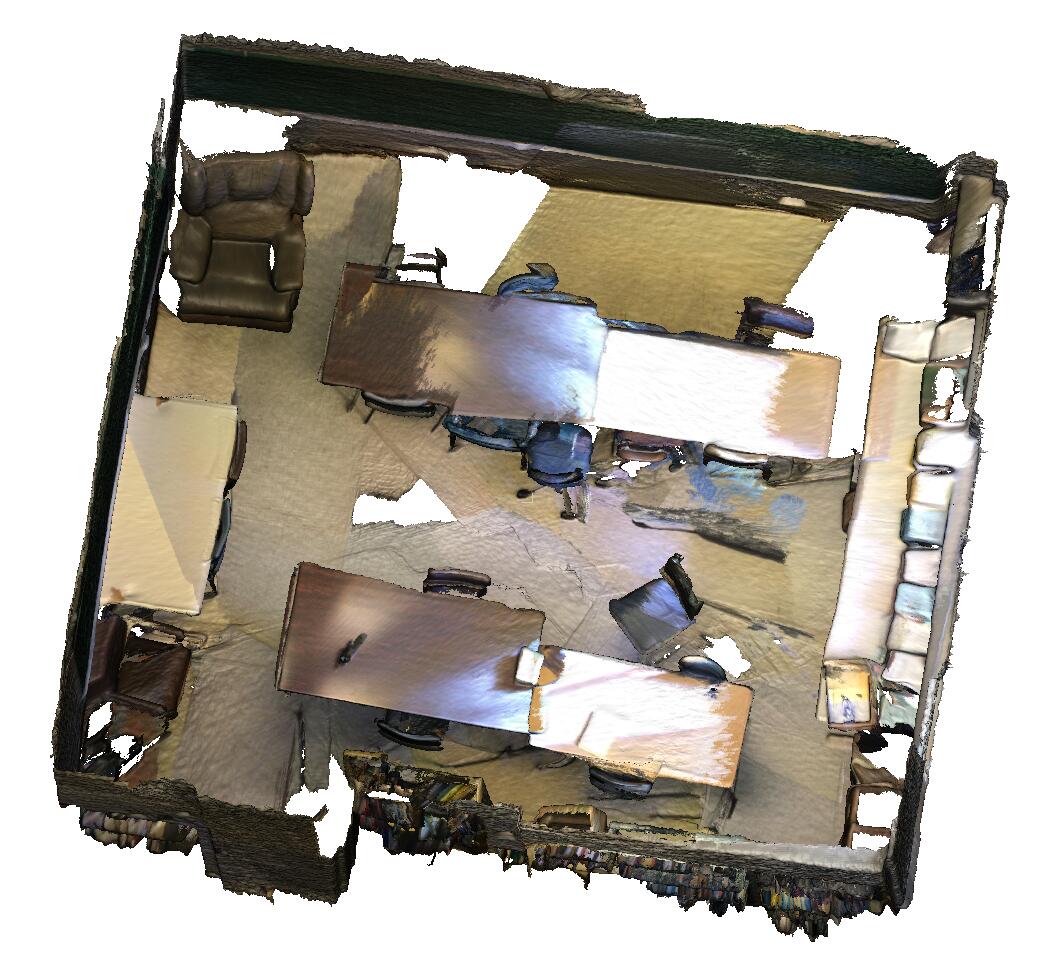}
  \caption{Inupt }
\end{subfigure}%
\begin{subfigure}{0.18\textwidth}
  \centering
  \includegraphics[width=\linewidth]{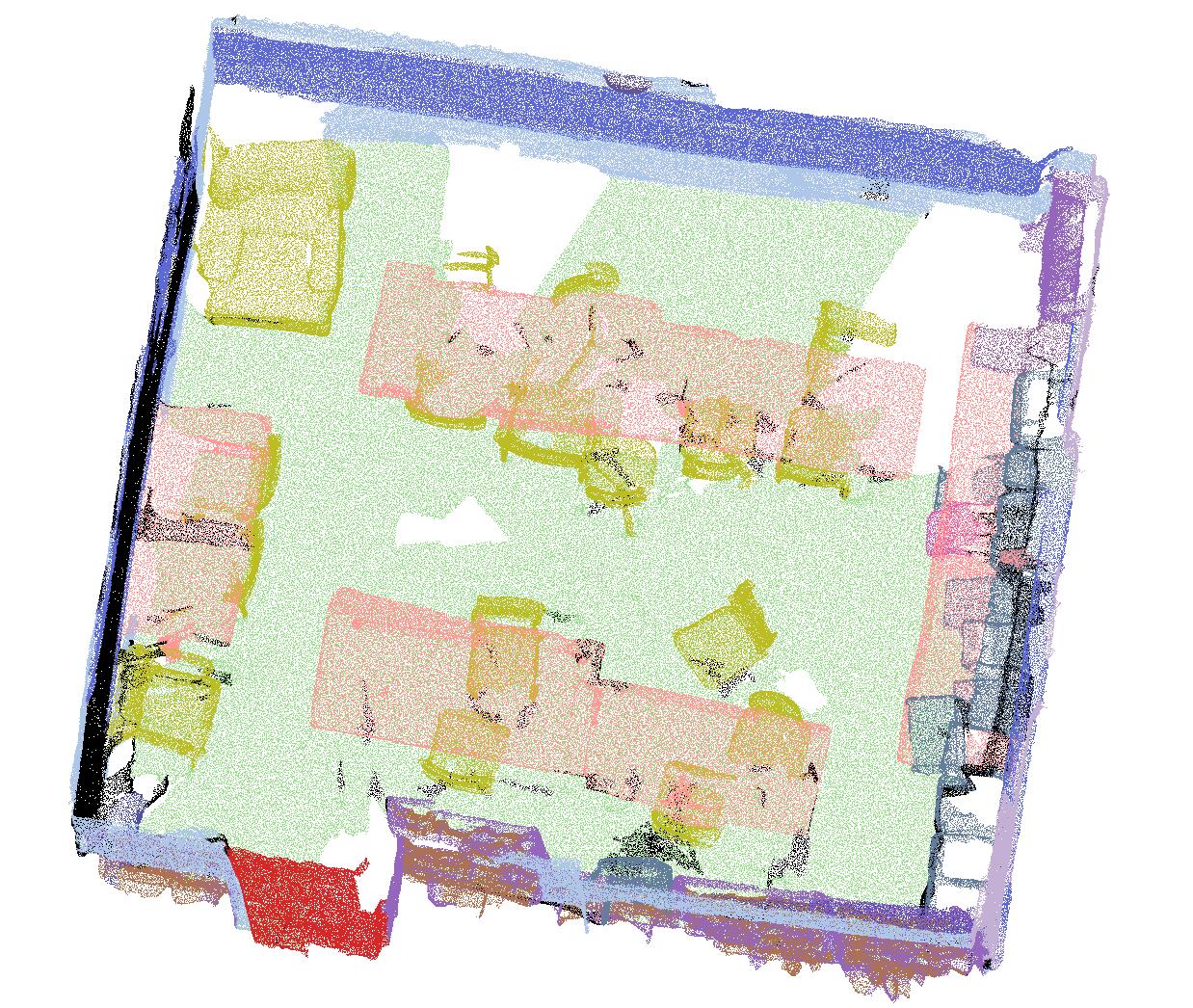}
  \caption{Semantic GT}
\end{subfigure}
\begin{subfigure}{0.18\textwidth}
  \centering
  \includegraphics[width=\linewidth]{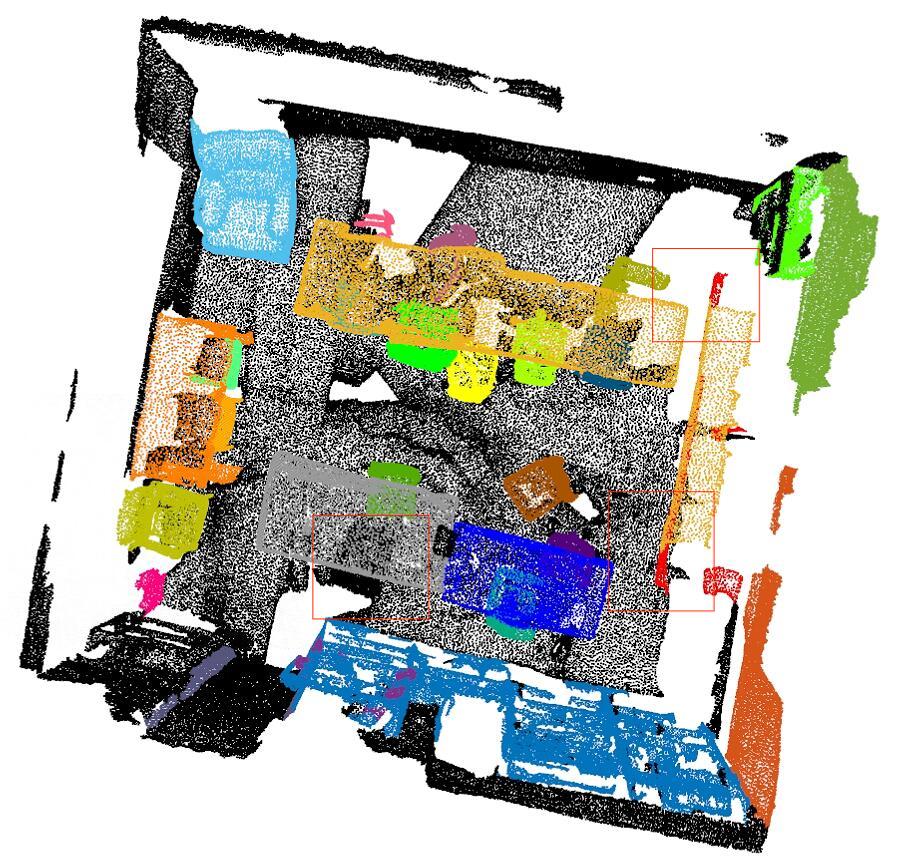}
  \caption{SSTnet}
\end{subfigure}
\begin{subfigure}{0.18\textwidth}
  \centering
  \includegraphics[width=\linewidth]{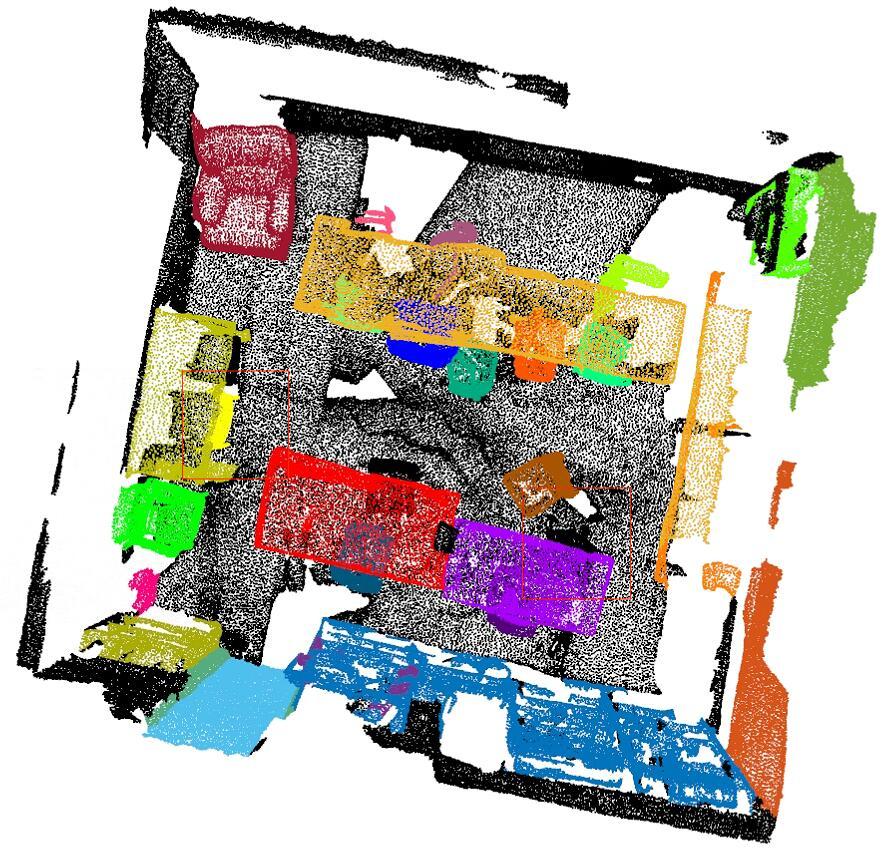}
  \caption{DKnet}
\end{subfigure}
\begin{subfigure}{0.18\textwidth}
  \centering
  \includegraphics[width=\linewidth]{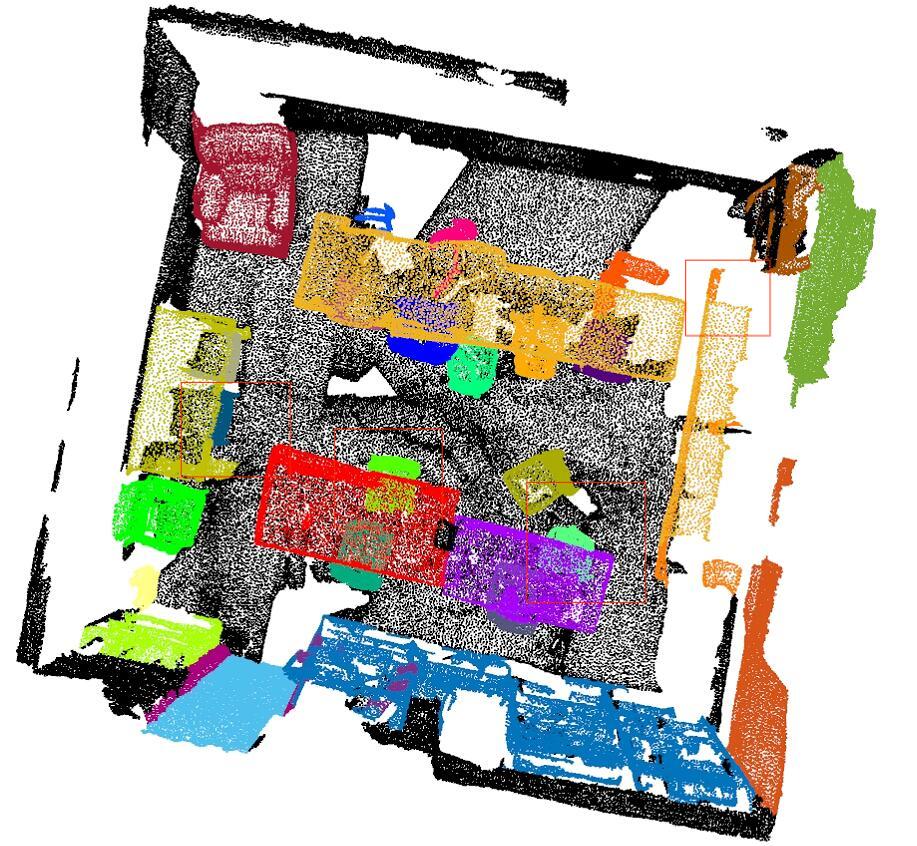}
  \caption{Ours}
\end{subfigure}
\caption{Visualiztion of instance segmentation results on the ScanNetv2 validation set.}
\label{fig:Visualiztion}
\end{figure*}

\textbf{Comparative qualitative results.} 
The visualization of 3D instance segmentation produced by our model, PSGfomer, is shown in Figure . \ref{fig:Visualiztion}. The results demonstrate that PSGformer is capable of correctly segmenting each instance, producing fine-grained segmentation results. The examples highlight the ability of our model to handle complex scenes, distinguishing individual instances clearly.

\begin{figure*}[pos=!h]
\centering
\begin{subfigure}{0.33\textwidth}
  \centering
  \includegraphics[width=\linewidth]{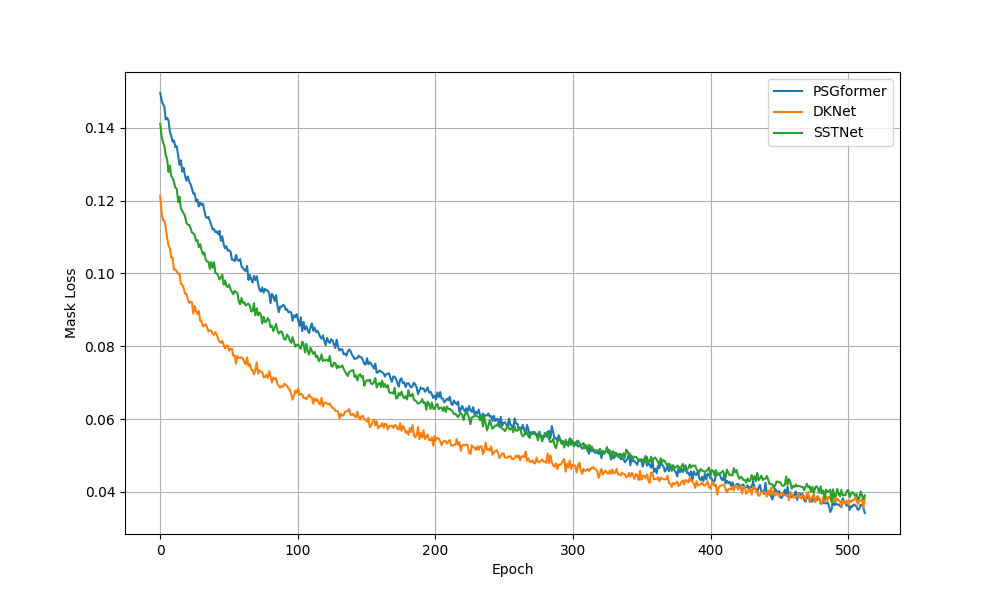}
  \caption{mask loss}
  \label{fig:sub1}
\end{subfigure}%
\begin{subfigure}{0.33\textwidth}
  \centering
  \includegraphics[width=\linewidth]{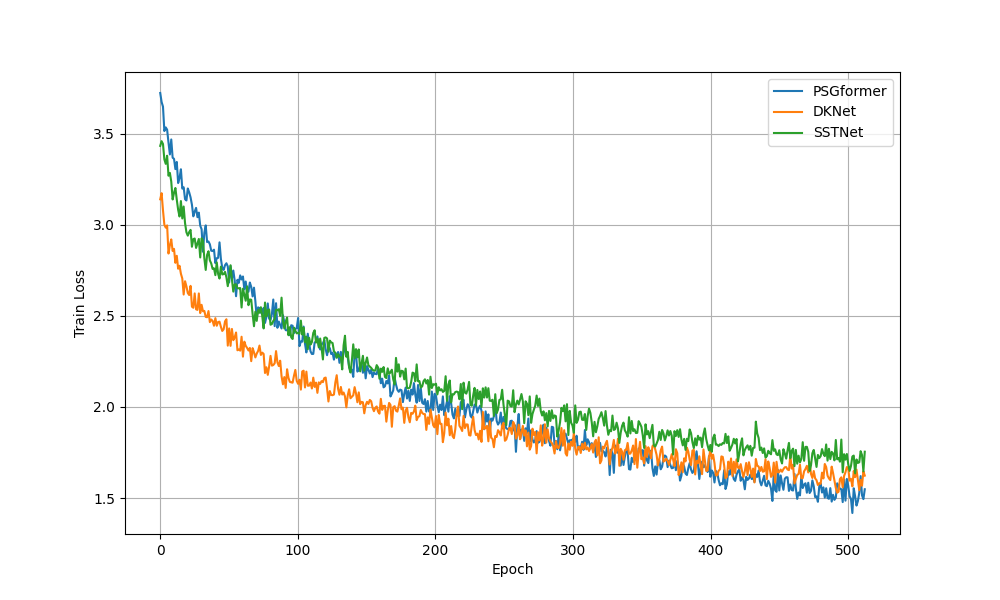}
  \caption{train loss}
\end{subfigure}
\begin{subfigure}{0.33\textwidth}
  \centering
  \includegraphics[width=\linewidth]{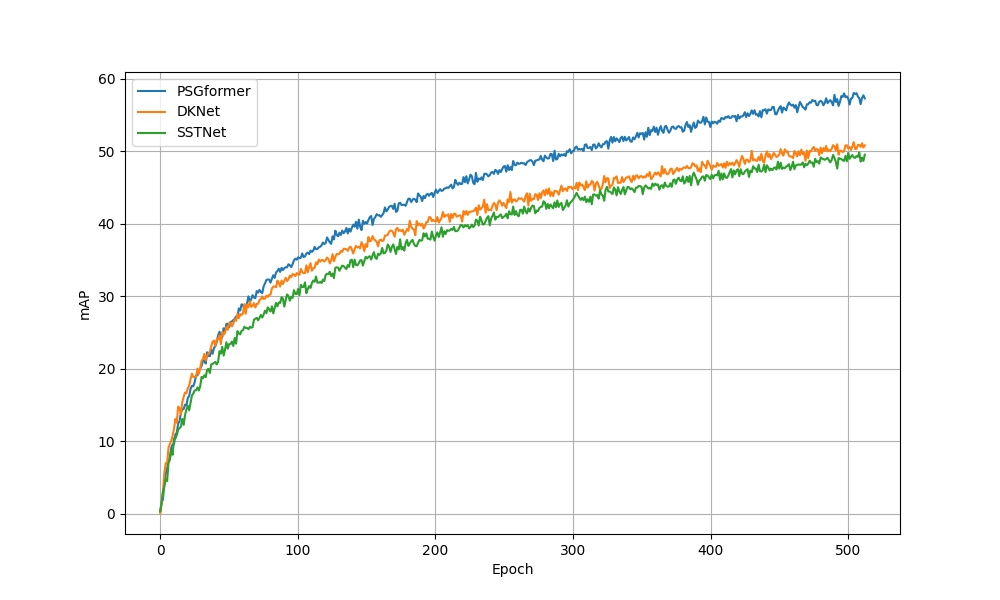}
  \caption{mAP}
\end{subfigure}
\caption{Comparison and visualization of loss curves of three kinds of networks during training.}
\label{fig:train loss}
\end{figure*}

\textbf{Comparative train loss.} 
Figure. \ref{fig:train loss} respectively illustrate the performance of PSGformer, DKNet, and SSTNet in terms of mask loss, train loss, and mean Average Precision (mAP) over 512 training epochs. Both mask loss and train loss demonstrate a trend of gradual increase with the advancement of training epochs, while mAP reflects the improvement of model performance as training progresses.

\begin{figure*}[pos=!h]
\centering

\begin{sideways}\textbf{Inupt}\end{sideways}\hfill
\begin{subfigure}{.18\textwidth}
\includegraphics[width=\linewidth]{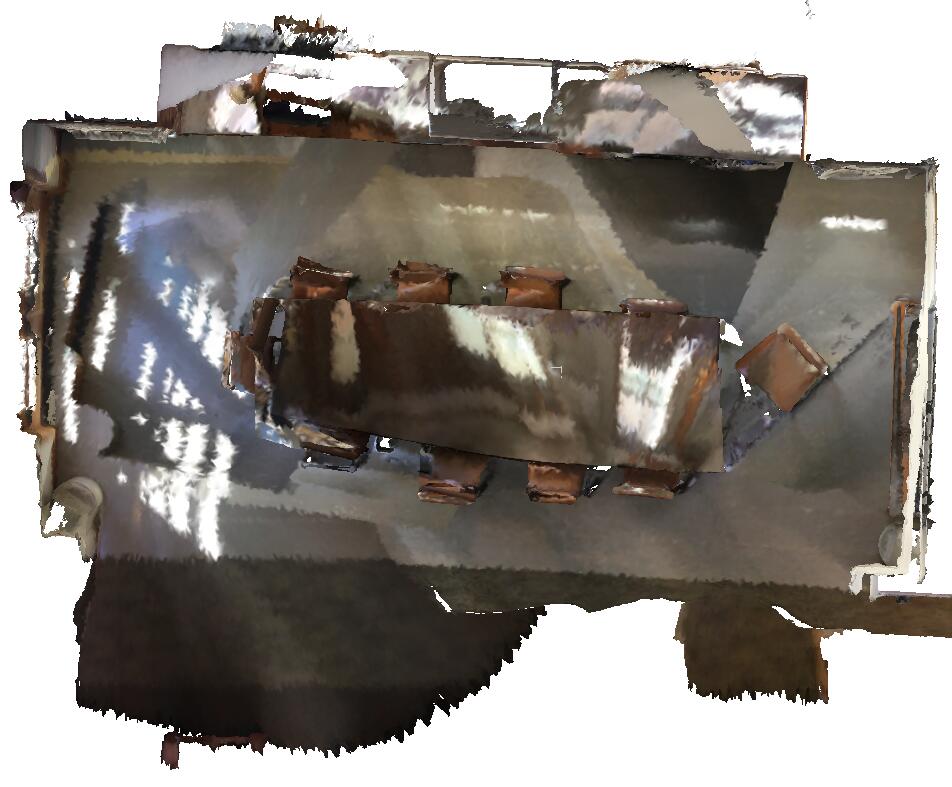}
\end{subfigure}\hfill
\begin{subfigure}{.18\textwidth}
\includegraphics[width=\linewidth]{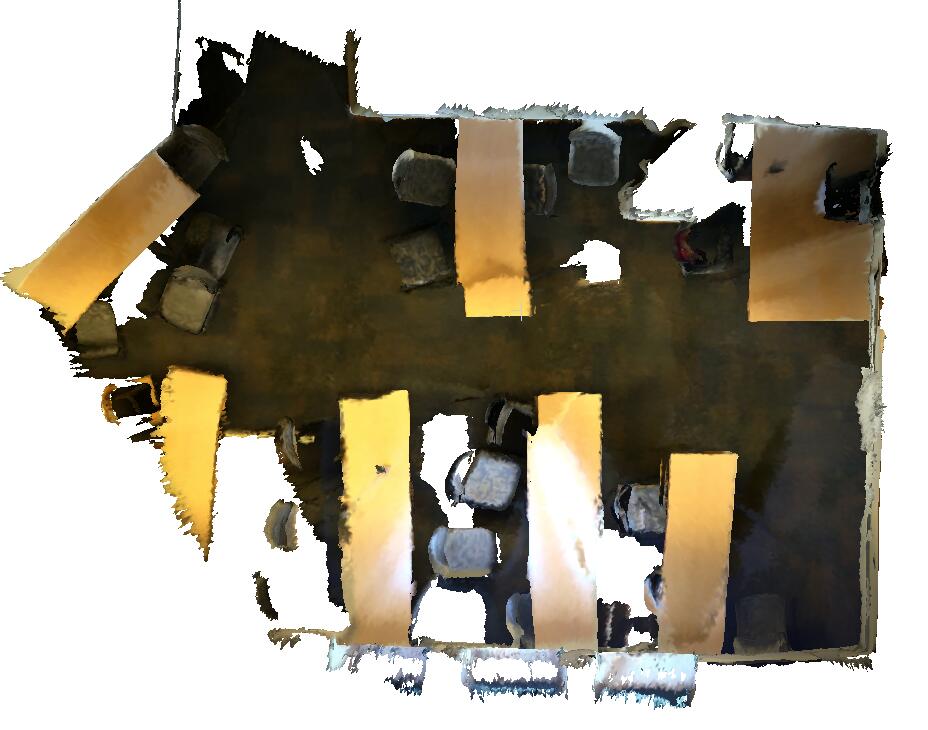}
\end{subfigure}\hfill
\begin{subfigure}{.18\textwidth}
\includegraphics[width=\linewidth]{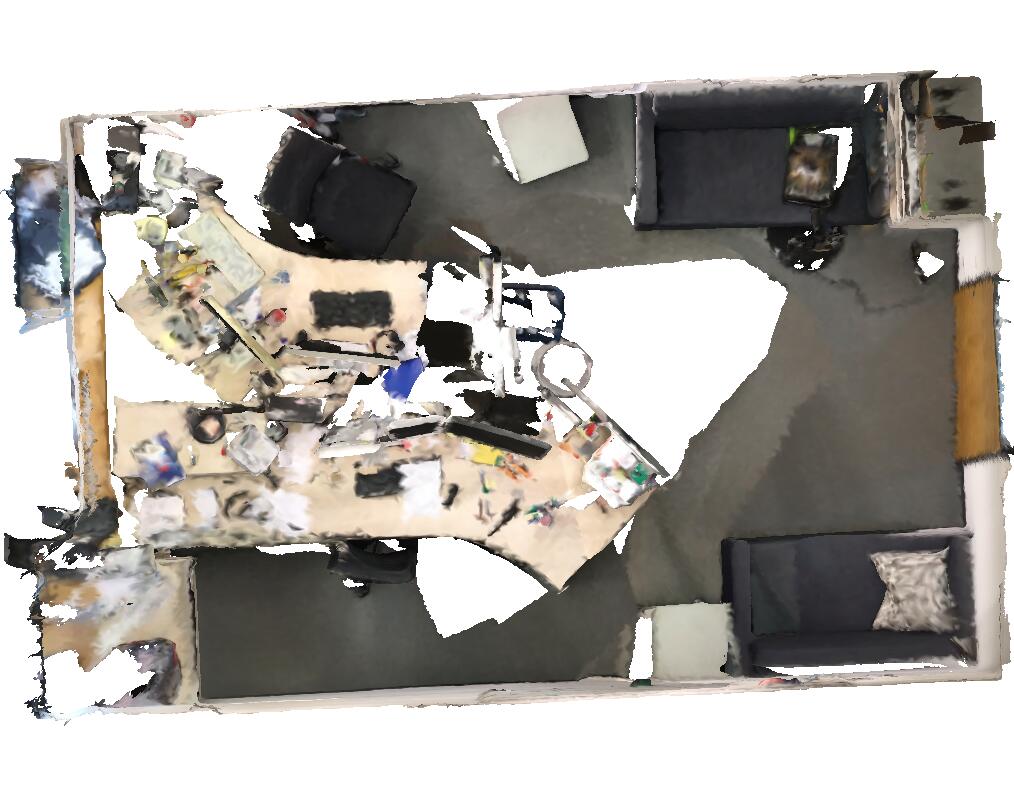}
\end{subfigure}\hfill
\begin{subfigure}{.18\textwidth}
\includegraphics[width=\linewidth]{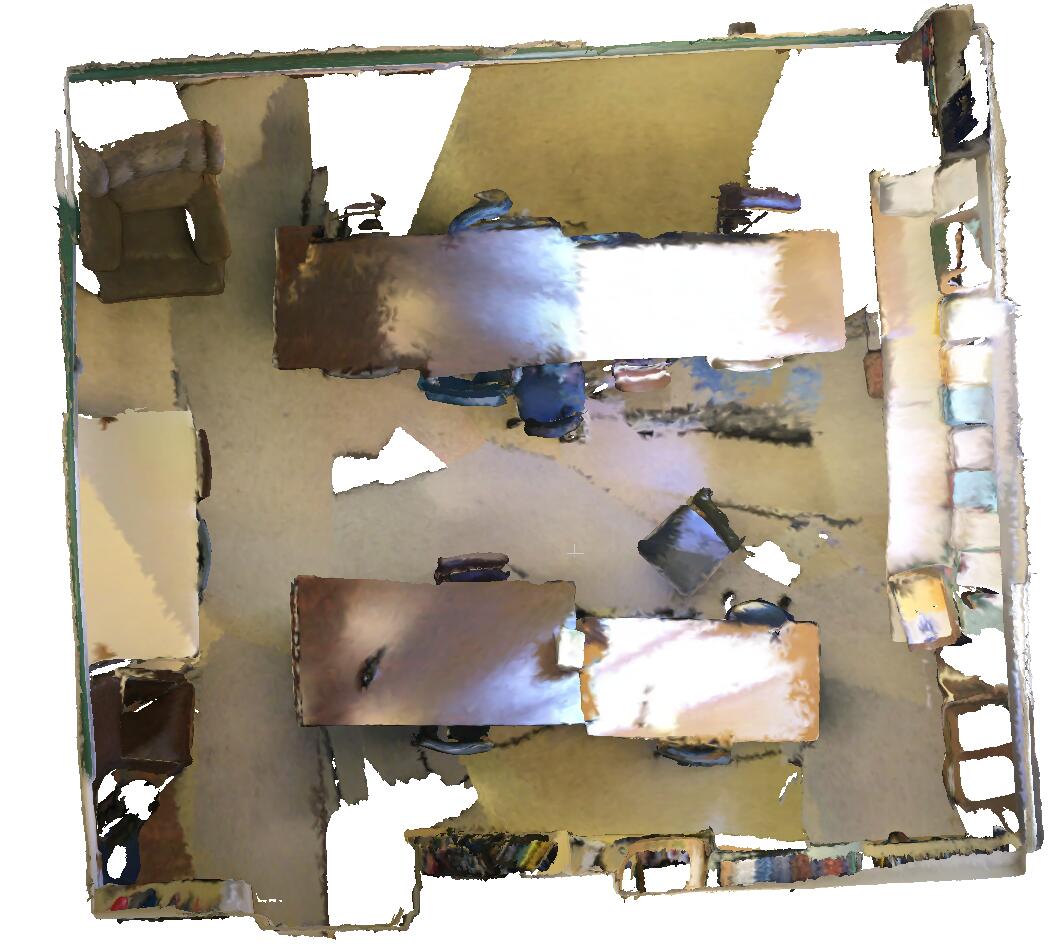}
\end{subfigure}\hfill
\begin{subfigure}{.18\textwidth}
\includegraphics[width=\linewidth]{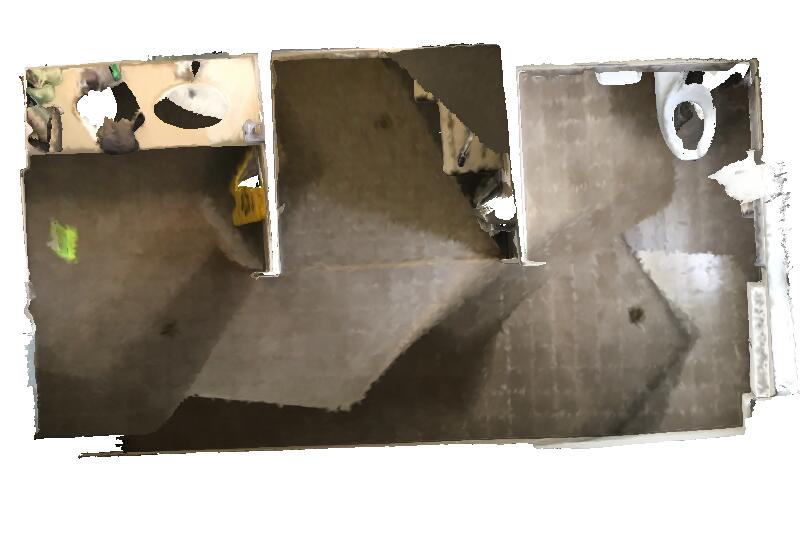}
\end{subfigure}

\medskip

\begin{sideways}\textbf{Semantic GT}\end{sideways}\hfill
\begin{subfigure}{.18\textwidth}
\includegraphics[width=\linewidth]{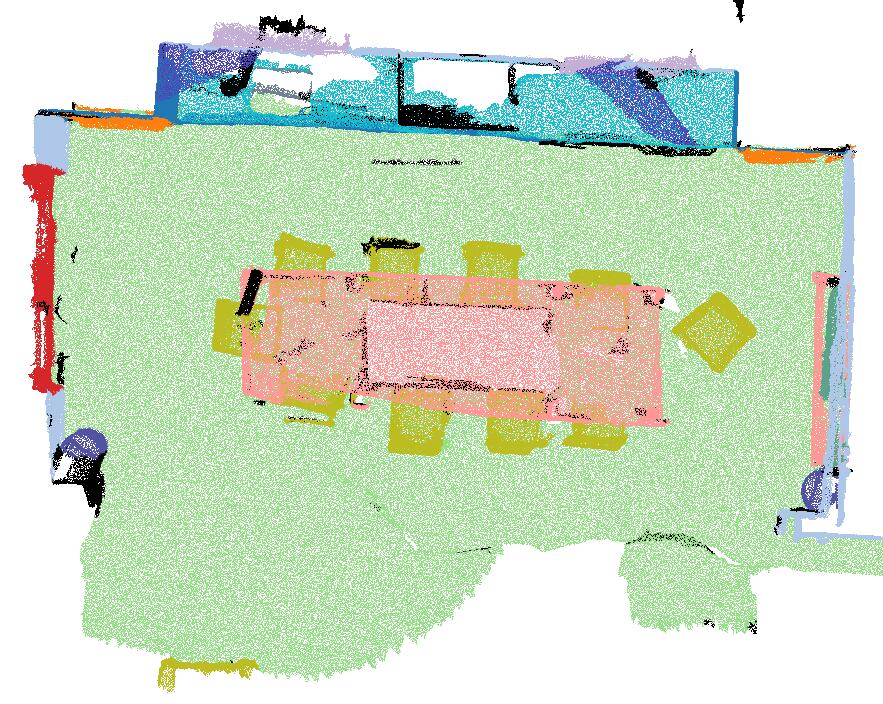}
\end{subfigure}\hfill
\begin{subfigure}{.18\textwidth}
\includegraphics[width=\linewidth]{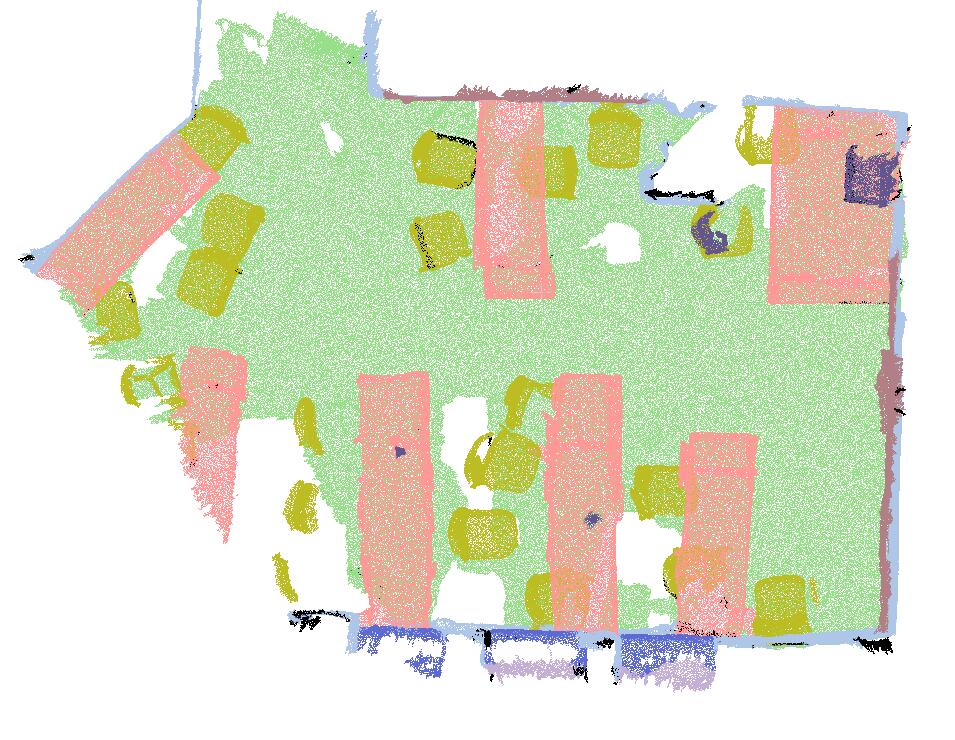}
\end{subfigure}\hfill
\begin{subfigure}{.18\textwidth}
\includegraphics[width=\linewidth]{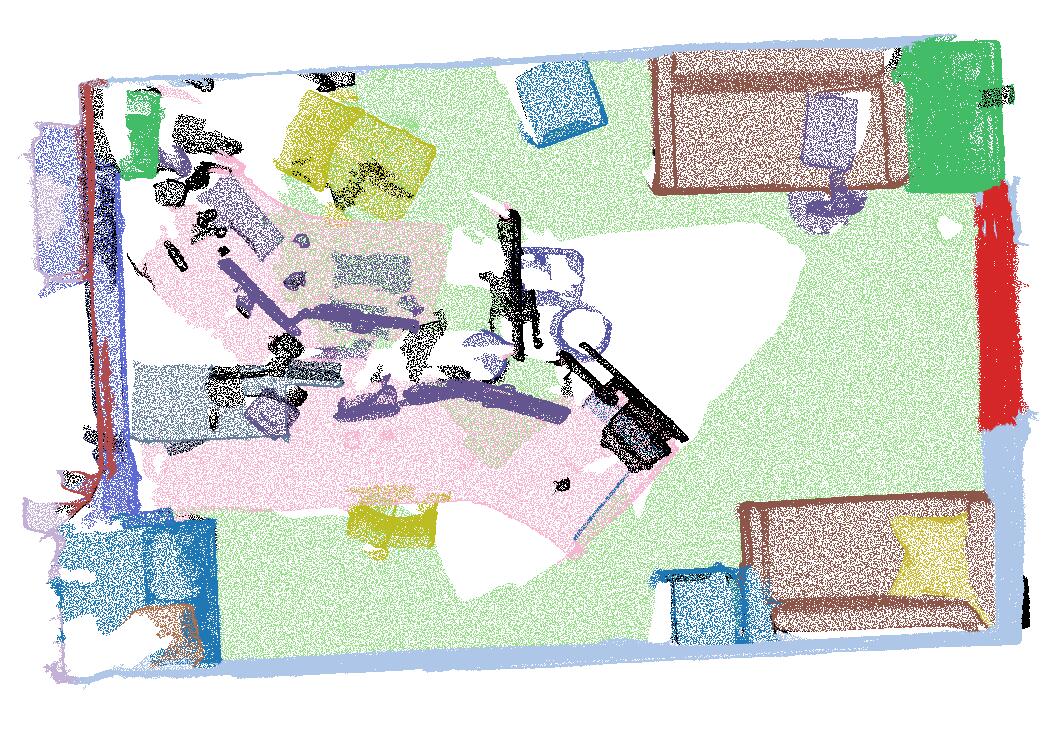}
\end{subfigure}\hfill
\begin{subfigure}{.18\textwidth}
\includegraphics[width=\linewidth]{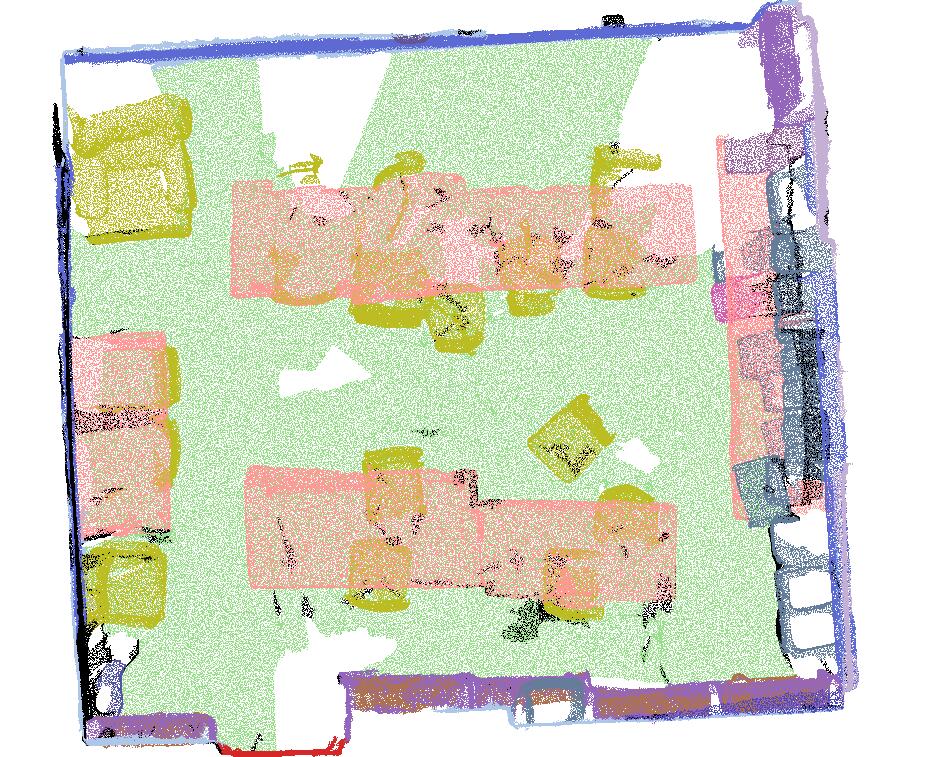}
\end{subfigure}\hfill
\begin{subfigure}{.18\textwidth}
\includegraphics[width=\linewidth]{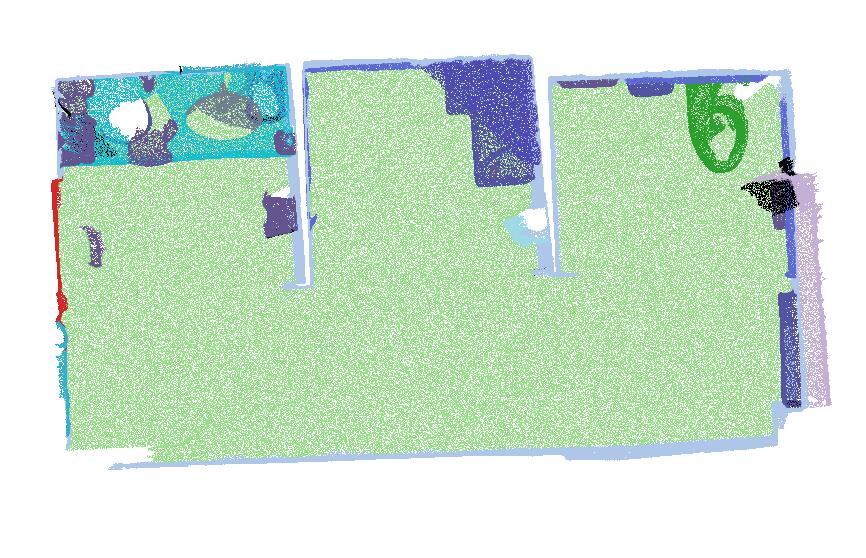}
\end{subfigure}

\begin{sideways}\textbf{Ours}\end{sideways}\hfill
\begin{subfigure}{.18\textwidth}
\includegraphics[width=\linewidth]{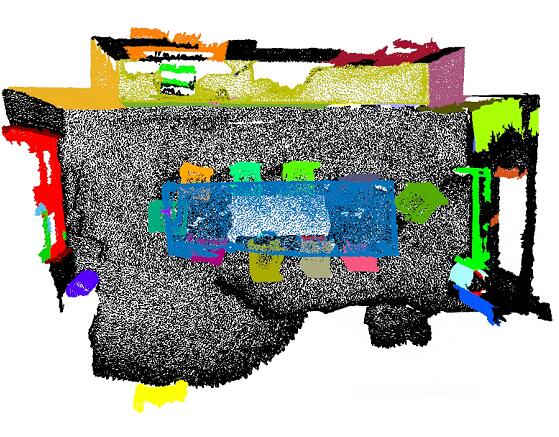}
\end{subfigure}\hfill
\begin{subfigure}{.18\textwidth}
\includegraphics[width=\linewidth]{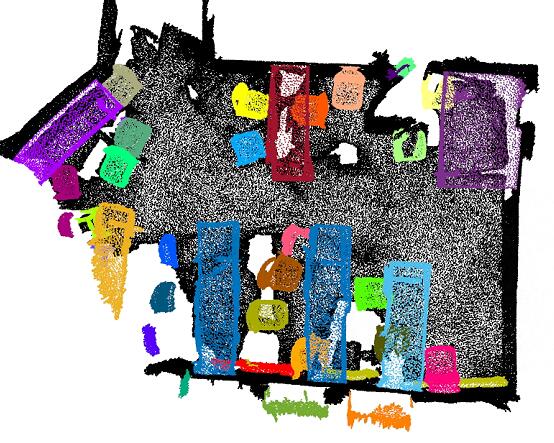}
\end{subfigure}\hfill
\begin{subfigure}{.18\textwidth}
\includegraphics[width=\linewidth]{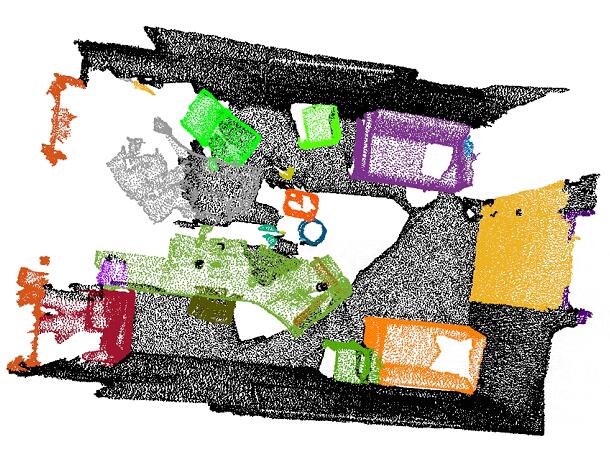}
\end{subfigure}\hfill
\begin{subfigure}{.18\textwidth}
\includegraphics[width=\linewidth]{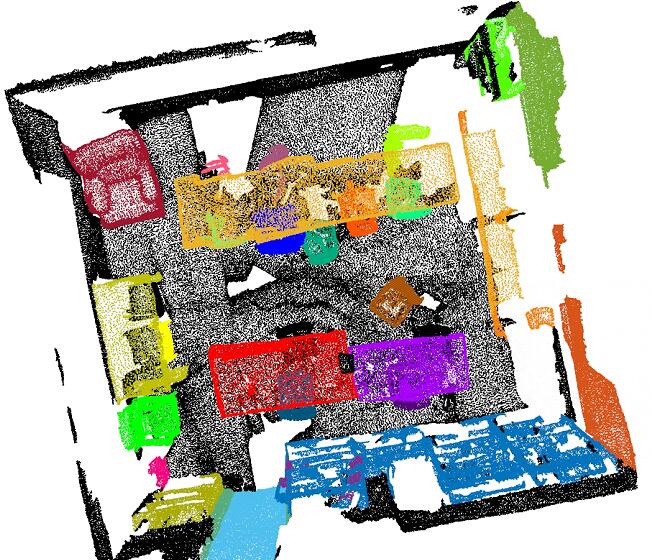}
\end{subfigure}\hfill
\begin{subfigure}{.18\textwidth}
\includegraphics[width=\linewidth]{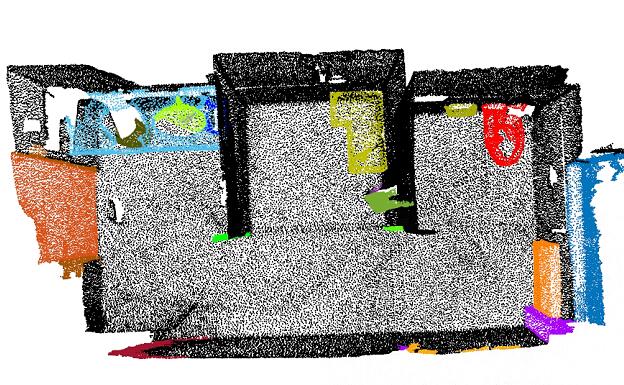}
\end{subfigure}

\caption{Qualitative results on ScanNetV2 dataset. Each column shows one example.}
\label{fig:Qualitative results}
\end{figure*}

\textbf{Qualitative Results of Our Approach}
The qualitative results of our approach on the ScanNetV2 are visualized in Figure. \ref{fig:Qualitative results}

\begin{figure*}[pos=!h]
\centering
\begin{subfigure}{0.24\textwidth}
  \centering
  \includegraphics[width=\linewidth]{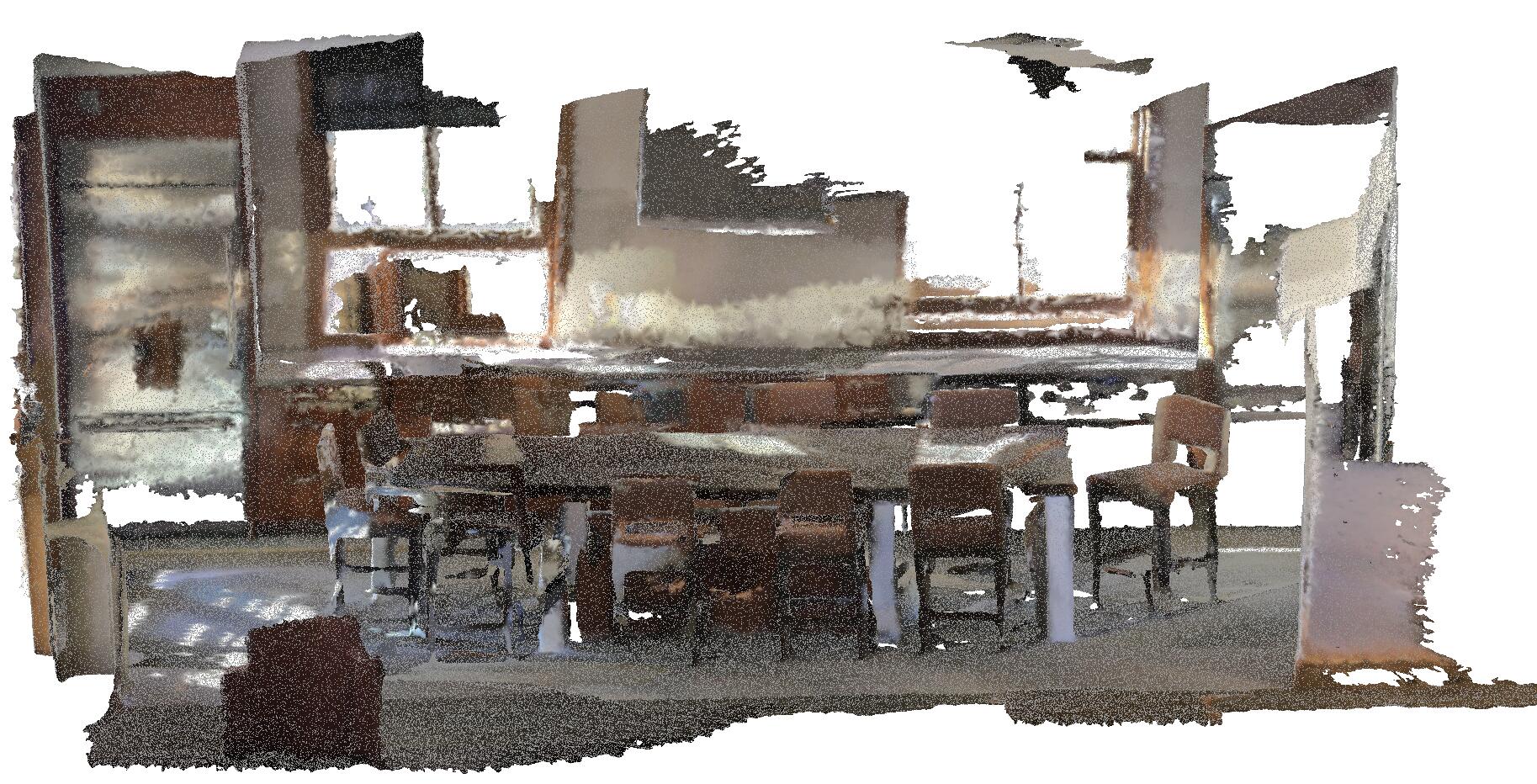}
  
  \label{fig:sub1}
\end{subfigure}%
\begin{subfigure}{0.24\textwidth}
  \centering
  \includegraphics[width=\linewidth]{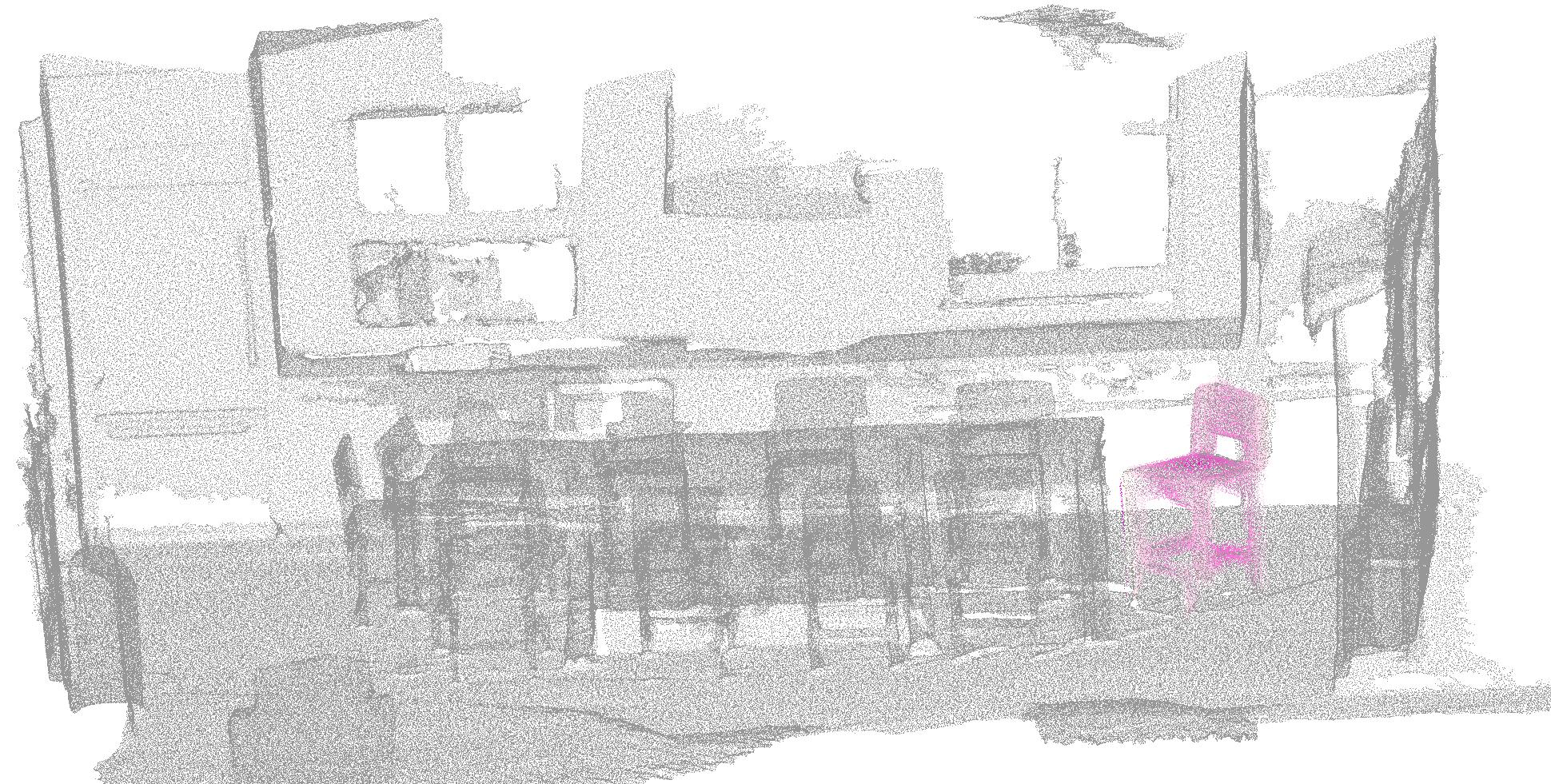}
  
  \label{fig:sub2}
\end{subfigure}
\begin{subfigure}{0.24\textwidth}
  \centering
  \includegraphics[width=\linewidth]{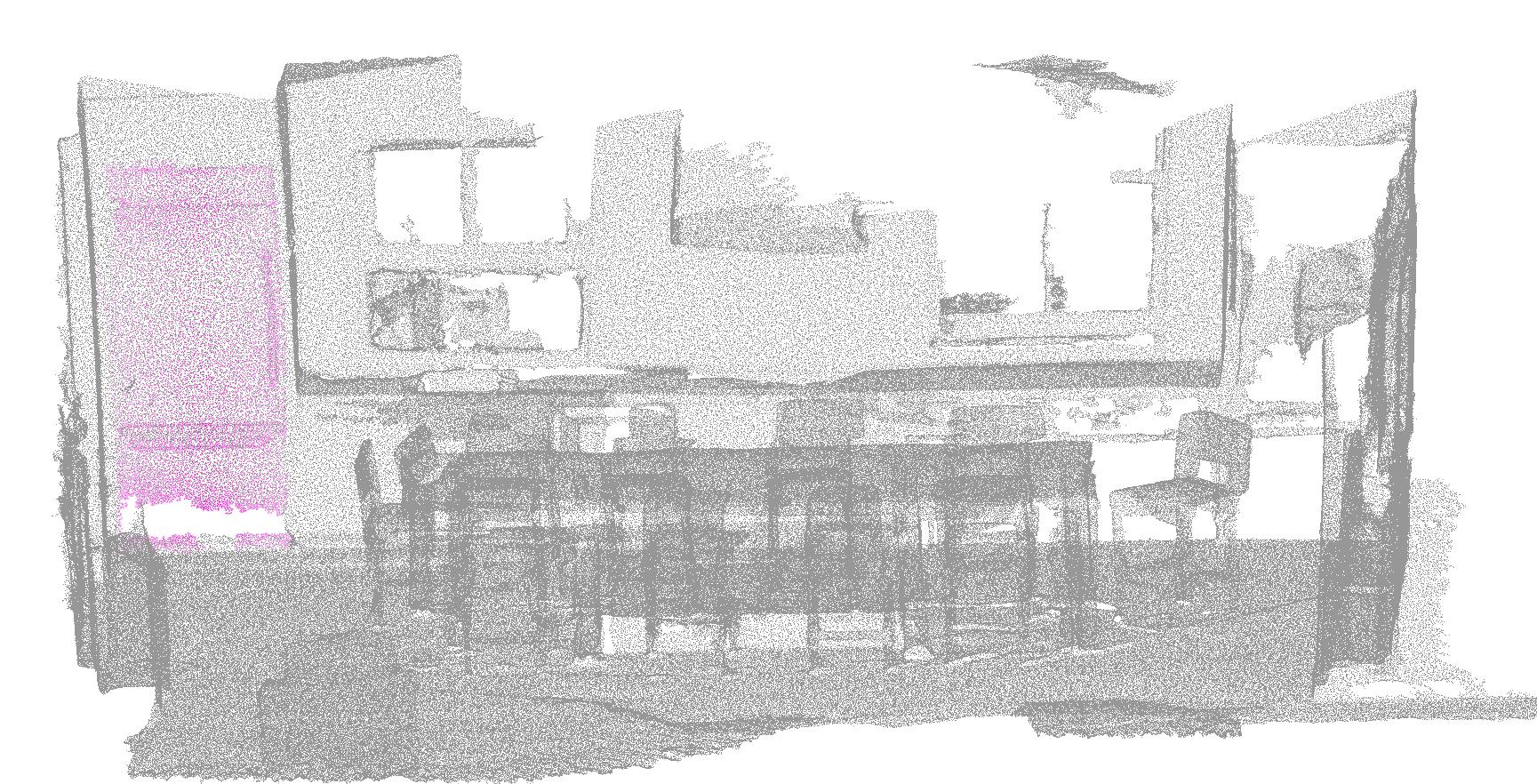}
  
  \label{fig:sub3}
\end{subfigure}
\begin{subfigure}{0.24\textwidth}
  \centering
  \includegraphics[width=\linewidth]{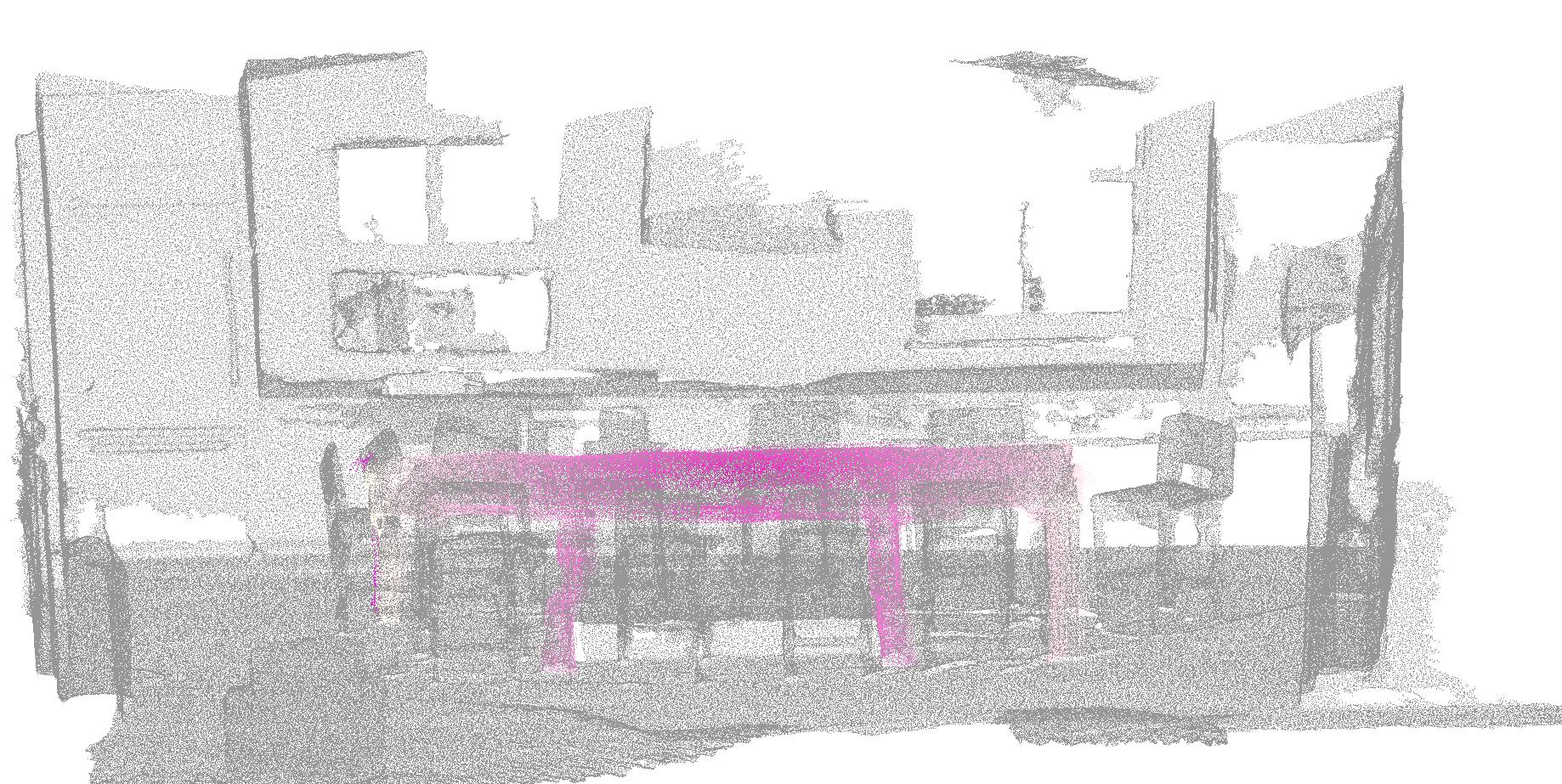}
  
  \label{fig:sub4}
\end{subfigure}
\caption{The visualization of the attention weight between the query vector and the corresponding real mask is shown, which corresponds to the mask prediction of different targets for each query vector in a scene.}
\label{fig:attention}
\end{figure*}

\textbf{Attention weight visualization.} 
Figure.\ref{fig:attention} visualizes the operation of the cross-attention mechanism within our PSGormer.  Given an input point cloud, the query vectors attend to superpoints, emphasizing regions of interest within the scene.  For visualization purposes, we propagate the attention weights of superpoints to their respective points.  The query vectors then carry this attention information forward, ultimately contributing to the final mask prediction generated by the prediction head.

\section{Conclusion}
In conclusion, we introduced the PSGformer, a groundbreaking 3D instance segmentation network that addresses the limitations of current models derived from semantic segmentation. The PSGformer represents a paradigm shift in the field, accurately predicting 3D instances by leveraging both global and local semantic information. Among its innovative aspects is the inclusion of a Multi-Level Semantic Aggregation Module, enabling the capture of nuanced scene features across different scales for a deeper comprehension of semantic information from both micro and macro perspectives. Additionally, our model independently processes super-point and aggregated features through the Parallel Feature Fusion Transformer Module, ensuring a more comprehensive feature representation and enhancing the feature learning process. Despite these advancements, our model has limitations that need addressing in future work, such as improving our Multi-Level Semantic Aggregation and Parallel Feature Fusion Modules to further augment the efficiency of 3D instance segmentation. The introduction of the PSGformer signifies a promising step forward in the field of 3D instance segmentation, providing new perspectives for future exploration.

\section{Acknowledgments}
This research was funded by Civil Aviation Flight University of China "Smart Civil Aviation" special project, grant number ZHMH2022-005. This research was funded by independent research project of civil aviation flight technology and flight safety key laboratory of Civil Aviation Flight University of China, grant number FZ2022ZZZ06. This research was supported by the Key Laboratory of Flight Techniques and Flight Safety, Civil Aviation Administratin of China , grant number FZ2022KF10
\bibliographystyle{elsarticle-num}
\bibliography{referees}

\end{document}